\documentclass[aps,prl,twocolumn,groupedaddress]{revtex4}
\usepackage{amsfonts}
\usepackage{mathrsfs}
\usepackage{graphicx}
\usepackage{subfigure}
\usepackage{amsmath}
\usepackage{amssymb}
\usepackage{amsbsy}
\usepackage{bm}
\usepackage{upgreek}
\usepackage{bbm}

\def\bw{\mathbf{w}}
\def\bx{\bm{\xi}}
\def\defe{\stackrel{{\rm def}}{=}}

\def\i{\mathrm{i}}

\def\by{\boldsymbol y}
\def\l{\left}
\def\r{\right}
\def\la{\left\langle}
\def\ra{\right\rangle}
\def\lame{\lambda_\epsilon}
\def\bxi{\boldsymbol \xi}
\def\bm{\boldsymbol m}
\def\qh{\hat{q}}
\def\rh{\hat{r}}

\def\mk{\mathcal{K}}

\begin{document}

\title{Weakly-correlated synapses promote dimension reduction in deep neural networks}
\author{Jianwen Zhou}
\author{Haiping Huang}
\affiliation{PMI Lab, School of Physics,
Sun Yat-sen University, Guangzhou 510275, People's Republic of China}
\date{\today}

\begin{abstract}
By controlling synaptic and neural correlations, deep learning has achieved empirical successes in improving classification performances. How synaptic correlations affect neural correlations to produce disentangled hidden representations remains elusive.
Here we propose a simplified model of dimension reduction, taking into account pairwise correlations among synapses, to reveal the mechanism underlying
how the synaptic correlations affect dimension reduction. Our theory determines the synaptic-correlation scaling form requiring only mathematical self-consistency, for both binary and continuous synapses. The theory also predicts that
weakly-correlated synapses encourage dimension reduction compared to their orthogonal counterparts. In addition, these synapses slow down the decorrelation process along the network depth. These two computational roles are explained by
the proposed mean-field equation. The theoretical predictions are in excellent agreement with numerical simulations, and the key features are also captured by a deep learning with Hebbian rules.
\end{abstract}

 \maketitle

\textit{Introduction.}---
Neural correlation is a common characteristic in most neural computations~\cite{Cohen-2011}, playing vital roles in stimulus coding~\cite{Coding-2016,NIPS-2010}, information storage~\cite{Brunel-2016} and various cognition tasks that can be
implemented by recurrent neural networks~\cite{Eric-2019,Neuron-2019}. Neural correlation was recently shown by a mean-field theory~\cite{Huang-2018} to be able to manipulate the dimensionality of layered representations in deep computations,
which was empirically revealed to be a fundamental process in deep artificial neural networks~\cite{Hinton-2006a}. This theory demonstrates that a compact neural representation with weak neural correlations is an emergent 
behavior of layered neural networks whose synaptic weights are \textit{independently and identically distributed}. However, in real cortical circuits, synaptic weights among neurons, even in the same layer,
may not be ideally independent with each other~\cite{Harris-2013,Harris-2015,Brunel-2016,CorreC-2009}, perhaps mainly due to biological synaptic plasticity~\cite{Brunel-2016,CorreC-2009}. On the other hand, a recent theoretical study of unsupervised 
feature learning predicts that weakly-correlated synapses promote unsupervised concept-formation by reducing the necessary sensory data samples~\cite{Huang-2019a,Huang-2019b}.
Therefore, in what exact way a weak correlation among synapses affects the emergent behavior of layered neural networks remains unknown.

In particular, suppression of unwanted variability in sensory inputs~\cite{Stef-2018}, compact representation of invariance (generalization to novel context)~\cite{Rust-2013},
and the feature selectivity (discrimination ability of neural representation)~\cite{DiCarlo-2007} may be closely related to an appropriate neural dimensionality, which keeps only relevant informative components spanning a robust neural subspace~\cite{Neuron-2019}.
Therefore, clarifying how synaptic correlations affect neural correlations and further the neural process 
underlying the transformation of representation dimensionality becomes fundamentally important.

\begin{figure}
    \includegraphics[bb=18 2 560 436,scale=0.35]{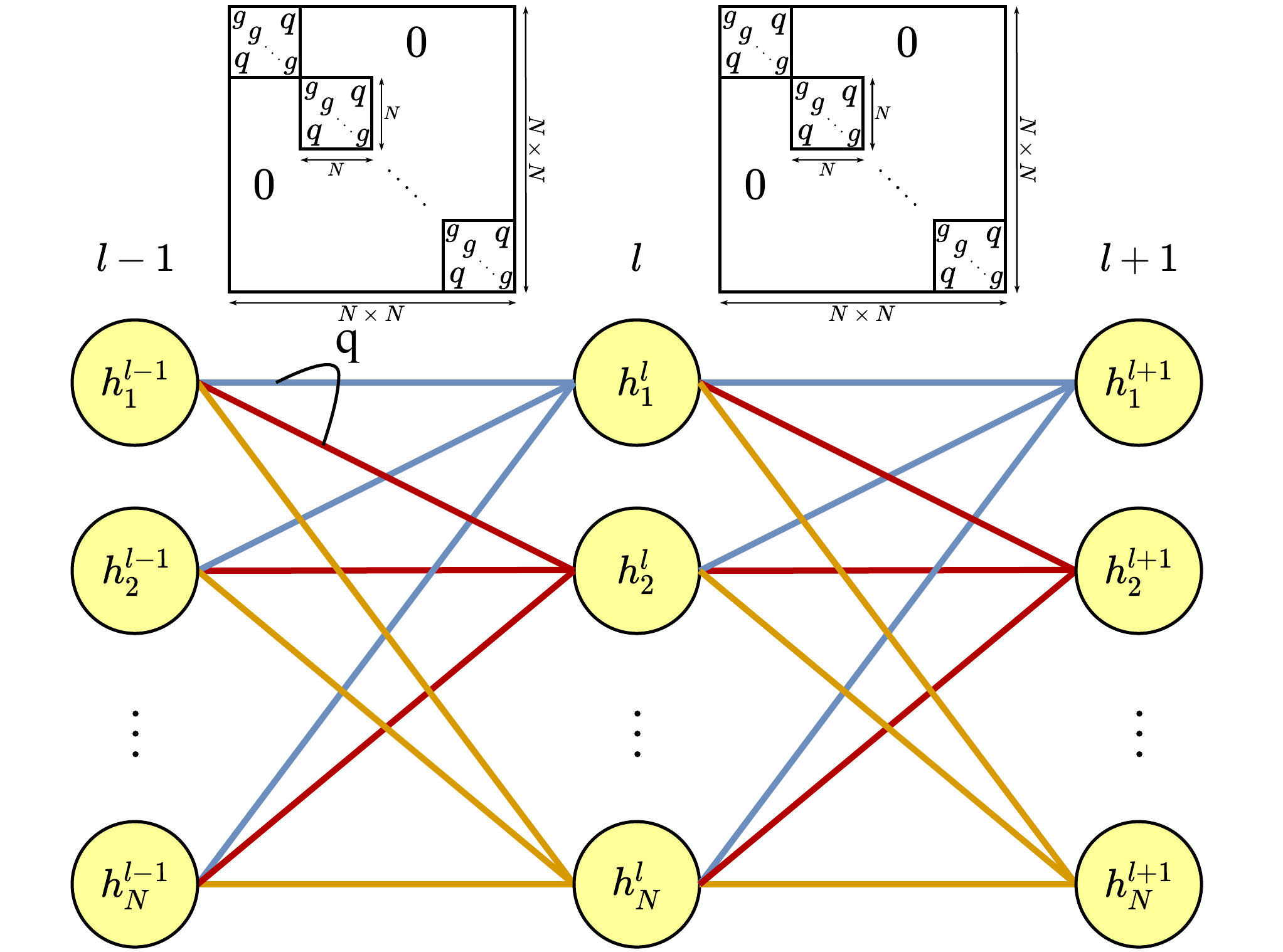}
  \caption{ (Color online) Schematic illustration of a deep neural network with correlated synapses. The deep neural network carries out a layer-wise transformation of a
  sensory input. During the transformation, a cascade of internal representations ($\{\mathbf{h}^l\}$) are generated by the correlated synapses, with the covariance structure specified by
  the matrix above the layer. $g$ characterizes the variance of synaptic weights, while the diagonal block characterizes the inter-receptive-field correlation among corresponding synapses (different line colors), 
  and $q$ specifies the synaptic
  correlation strength. We do not know \textit{a priori} the exact scaling form of $q$, which is self-consistently determined by our theory.
  }\label{dnn}
\end{figure}

To reveal the mechanism underlying how the synaptic correlations affect compact neural representations, we consider deep neural networks that realize a layer-wise transformation of sensory inputs. All in-coming synapses to a hidden neuron form a receptive field (RF) of that hidden neuron. 
The correlation among synapses is modeled by
the inter-RF correlation (Fig.~\ref{dnn}). We do not need a prior knowledge about the synaptic correlation strength. In fact, our mean-field theory yields different scaling behaviors of synaptic correlation with respect to
the number of neurons at each layer, for both binary and continuous synaptic weights. The scaling behaviors are exactly a requirement of mathematically well-defined dimensionality.

Compared to the previous work of orthogonal weights~\cite{Huang-2018}, our current model reveals richer ways of controlling dimensionality of neural representations, i.e., the dimensionality can be tuned layer by layer in both additive and multiplicative manners.
Each manner can be analytically understood in the mean-field limit. The theory predicts that, according to the scaling, the weak correlation among inter-RF synapses is able to promote dimension reduction in deep neural networks. In addition, these synapses slow down the neural decorrelation process
along the network hierarchy. Therefore, our theory provides deep insights towards understanding how synaptic correlations shape compact neural representations in deep neural networks.

\textit{Model.}---
A deep neural network is composed of multiple layers of non-linear transformation of sensory inputs (e.g., natural images). The depth of the network is defined as
the number of hidden layers ($d$), and the width of each layer is defined by the number of neurons at that layer. For simplicity, we assume an equal width ($N$) in this study.
To specify weights between $l-1$ and $l$-th layers, we define a weight matrix $\mathbf{w}^l$ 
whose $i$-th row corresponds to incoming connections to the neuron $i$ at the higher layer (so-called the receptive field of the neuron $i$).
Firing biases of neurons at the $l$-th
layer are denoted by $\mathbf{b}^l$. The input data are transformed through consecutive hidden representations denoted by  
$\mathbf{h}^l$ ($l=1,\cdots,d$), in which each entry $h_i^l$ defines a non-linear transformation of its pre-activation $z_i^l\equiv[\mathbf{w}^l\mathbf{h}^{l-1}]_i+b_i^l$, as
$h_i^l=\phi(z_i^l)$. Without loss of generality, we use the non-linear transfer
function $\phi(x)=\tanh(x)$.

To take into account inter-RF correlations, we specify the covariance structure as $\overline{w_{ik}^lw_{jk}^l}=q$ for $i\neq j$, and $\overline{(w_{ik}^l)^2}=g^2/N$ for continuous weights, while 
$\overline{(w_{ik}^l)^2}=1$ for binary weights ($\pm1$) (see Fig.~\ref{dnn}). The weight has zero mean. The bias follows $\mathcal{N}(0,\sigma_b)$. The orthogonal RF case was studied in a recent work~\cite{Huang-2018} that demonstrates
mechanisms of dimension reduction in deep neural networks of continuous weights. Here, we extend the analysis to the non-orthogonal case, and do not enforce any 
scaling constraint \textit{a priori} to the correlation level $q$. The role of $q$ in dimension reduction is determined in a self-consistent way. Note that we do not assume any prescribed
correlations of intra-RF weights for simplicity. In the current setting, the weight matrix $\bw^l$ is more highly structured than in the orthogonal case, resembling qualitatively what occurs in a biological neural circuit~\cite{Harris-2015,CorreC-2009}.

To define the computational task, we consider a random input ensemble
from which each input sample is drawn. This ensemble is characterized by zero mean and the covariance matrix $\Lambda=\frac{1}{N}\bx\bx^{{\rm T}}$, where $\bx$ is an $N\times P$ matrix whose
components follow a normal distribution of zero mean and variance $\sigma^2$ ($\sigma=0.5$ throughout the paper). The ratio $\alpha=P/N$ controls the spectral density of the covariance matrix~\cite{SM}.

We first analyze the continuous-weight case.
Considering an average over the input ensemble, we define the mean-subtracted weighted-sum as $a_i^l=z_i^l-\left<z_i^l\right>=\sum_jw^l_{ij}(h^{l-1}_j-\left<h_j^{l-1}\right>)$, and thus $a_i^l$ has zero mean.
It follows that the covariance of $\mathbf{a}^l$ can be written as $\Delta^l_{ij}=\left<a_i^la_j^l\right>=\left[\mathbf{w}^l\mathbf{C}^{l-1}(\mathbf{w}^l)^{{\rm T}}\right]_{ij}$, where $\mathbf{C}^{l-1}$
defines the covariance (two-point correlation) matrix of the hidden representation at the $(l-1)$-th layer. The deep network defined in Fig.~\ref{dnn} implies that 
each neuron at an intermediate layer receives a large number of nearly independent input contributions. Therefore, 
the central limit theorem suggests that the mean of hidden neural activity 
$\mathbf{m}^l$ and covariance $\mathbf{C}^l$ are given respectively by
\begin{subequations}\label{dMFT}
\begin{align}
 m_i^l&=\left<h_i^l\right>=\int Dt\phi\left(\sqrt{\Delta^l_{ii}}t+[\mathbf{w}^l\mathbf{m}^{l-1}]_i+b_i^l\right),\\
 \begin{split}
 C^l_{ij}&=\int DxDy\phi\left(\sqrt{\Delta_{ii}^l}x+b_i^l+[\mathbf{w}^l\mathbf{m}^{l-1}]_i\right)\phi\left(\sqrt{\Delta_{jj}^l}\right.\\
 &\left. (\uppsi x+y\sqrt{1-\uppsi^2})+b_j^l+[\mathbf{w}^l\mathbf{m}^{l-1}]_j\right)-m_i^lm_j^l,
 \end{split}
 \end{align}
\end{subequations}
where $Dx\equiv\frac{e^{-x^2/2}dx}{\sqrt{2\pi}}$, and $\uppsi\equiv\frac{\Delta_{ij}^l}{\sqrt{\Delta_{ii}^l\Delta_{jj}^l}}$. To derive Eq.~(\ref{dMFT}), we re-parametrize $a_i^l$ and $a_j^l$ by independent normal random variables, such that the 
covariance structure of the mean-subtracted pre-activations is satisfied~\cite{SM}. In physics, Eq.~(\ref{dMFT}) constructs 
an iterative mean-field equation across layers to
describe the transformation of the activity statistics in the deep neural hierarchy. 

In statistical physics, the macroscopic behavior of a complex system of many degrees of freedom can be described by a few order parameters. Following the same spirit,
we define a linear dimensionality of the hidden representation at each layer as
\begin{equation}\label{dim}
 D^l=\frac{\left(\sum_{i=1}^N\lambda_i\right)^2}{\sum_{i=1}^N\lambda_i^2},
\end{equation}
where $\{\lambda_i\}$ is the eigen-spectrum of the covariance matrix $\mathbf{C}^l$. In statistics, this measure is called
the participation ratio~\cite{NIPS-2010} that is used to identify 
the number of non-zero significant eigenvalues. These eigenvalues capture the dominant dimensions that explain variability of the neural representation.

Using the mathematical identities ${\rm tr}(\mathbf{C})=\sum_i\lambda_i$ and ${\rm tr}(\mathbf{C}^2)=\sum_i\lambda_i^2$,
one can derive that the normalized dimensionality $\tilde{D}^l\equiv D^l/N=\frac{(\mathcal{K}_1^l)^2}{N\Upsigma^l+\mathcal{K}_2^l}$, where
$\Upsigma^l=\frac{2}{N^2}\sum_{i<j}(C_{ij}^l)^2$, $\mathcal{K}_1^l\equiv\frac{1}{N}\sum_iC_{ii}^l$, and $\mathcal{K}_2^l\equiv\frac{1}{N}\sum_i(C_{ii}^l)^2$.
$\Upsigma^l$ characterizes the overall neural correlation strength. In a mean-field approximation~\cite{Mezard-1987}, $C_{ij}^l\sim\mathcal{O}(1/\sqrt{N})$, which implies that
$\Delta_{ij}^l$ is of the same order
$\mathcal{O}(1/\sqrt{N})$. Therefore, $C_{ij}^l$ can be expanded in terms of $\Delta_{ij}^l$, resulting in $C_{ij}^l=K_{ij}^l\Delta_{ij}^l$ to leading order~\cite{SM}.
Here, $K_{ij}^l\defe\int Dx\phi'(g\sqrt{\mathcal{K}_1^{l-1}}x+z_i^0)\int Dy\phi'(g\sqrt{\mathcal{K}_1^{l-1}}y+z_j^0)$ where the mean $z_{i,j}^0=b_{i,j}^l+[\bw^l\mathbf{m}^{l-1}]_{i,j}$.

For the binary-weight case, the pre-activation should be multiplied by a pre-factor $\frac{g}{\sqrt{N}}$ to ensure that the pre-activation is of the order one.
By using the above expansion, one gets the relationship between $\Upsigma^{l+1}$ and $\Upsigma^{l}$ as
\begin{equation}\label{bsgm}
 N\Upsigma^{l+1}=g^4\kappa\left[(1+q^2)N\Upsigma^{l}+(1-q^2)\mathcal{K}_2^l+q^2N(\mathcal{K}_1^l)^2\right],
\end{equation}
where the overline in $\kappa\equiv\overline{(K_{ij}^{l+1})^2}$ means the disorder average over the quenched network parameters~\cite{SM}. Clearly, $q$ can not be
of the order one, otherwise Eq.~(\ref{bsgm}) is not self-consistent in mathematics, in that $N\Upsigma\sim\mathcal{O}(1)$ because of the magnitude of $C_{ij}$.
A unique scaling for $q$ must then be $q=\frac{r}{\sqrt{N}}$, resulting in $q^2N=r^2$ where $r\sim\mathcal{O}(1)$, and thus Eq.~(\ref{bsgm}) is self-consistent in physics as well. 
We thus call this type of synapses the weakly-correlated synapses.
Inserting $\Upsigma^{l+1}$
into the dimensionality definition together with the linear approximation of $C_{ii}^{l+1}\simeq K_{ii}^{l+1}\Delta_{ii}^{l+1}=g^2K_{ii}^{l+1}\mathcal{K}_1^l$~\cite{SM}, one immediately obtains 
the dimensionality of the hidden representation at the ($l+1$)-th layer, in terms of the activity statistics from the previous layer,
\begin{equation}\label{bdim}
 \tilde{D}^{l+1}=\frac{(\mathcal{K}_1^l)^2}{\upgamma_1\left(N\Upsigma^l+\mathcal{K}_2^l\right)+(\upgamma_1r^2+\upgamma_2)(\mathcal{K}_1^l)^2},
\end{equation}
where $\upgamma_1\equiv\frac{\overline{(K_{ij}^{l+1})^2}}{\overline{K_{ii}^{l+1}}^2}$, and $\upgamma_2\equiv\frac{\overline{(K_{ii}^{l+1})^2}}{\overline{K_{ii}^{l+1}}^2}$. Using the Cauchy-Schwartz inequality, one can prove that $\upgamma_1\leq\upgamma_2$~\cite{SM}.
It is also clear that $\upgamma_2\geq1$.

For the continuous-weight case, following the same line of derivation as above, one obtains a similar relationship between $\Upsigma^{l+1}$ and $\Upsigma^{l}$ as~\cite{SM}
\begin{equation}\label{csgm}
 N\Upsigma^{l+1}=\kappa\left[(g^4+N^2q^2)(N\Upsigma^{l}+\mathcal{K}_2^l)+q^2N^3(\mathcal{K}_1^l)^2\right].
\end{equation}
The self-consistency of the above equation for the neural correlation strength requires that $q=\frac{r}{N^{3/2}}$ where $r\sim\mathcal{O}(1)$. Therefore, $q^2N^2$ vanishes in a large network width limit.
It then follows that the output dimensionality given the input activity statistics reads
\begin{equation}\label{cdim}
 \tilde{D}^{l+1}=\frac{(\mathcal{K}_1^l)^2}{\upgamma_1\left(N\Upsigma^l+\mathcal{K}_2^l\right)+(\upgamma_1r^2/g^4+\upgamma_2)(\mathcal{K}_1^l)^2}.
\end{equation}
When $q=0$, $\upgamma_1=1$, and the dimensionality formula for the orthogonal case is recovered~\cite{Huang-2018}. Note that the dimensionality of the previous hidden representation for both binary- and continuous-weight cases is given by
$\tilde{D}^l=\frac{(\mathcal{K}_1^l)^2}{N\Upsigma^l+\mathcal{K}_2^l}$. Therefore, the output dimensionality is tuned by a multiplicative factor $\upgamma_1$ and an additive
term [the last term in the denominator of Eq.~(\ref{bdim}) or Eq.~(\ref{cdim})]. The tuning mechanism of the hidden-representation
dimensionality in
a deep hierarchy is thus much richer than that in the orthogonal scenario~\cite{Huang-2018}.

The linear relationship between $\Upsigma^{l+1}$ and $\Upsigma^l$ in Eq.~(\ref{bsgm}) and Eq.~(\ref{csgm}) defines an operating point where $\Upsigma^{l+1}=\Upsigma^l$.
When the input neural correlation strength is below the point, the non-linear transformation would further strengthen the correlation; whereas, the output neural correlation would be attenuated
once the input one is above the operating point. The synaptic correlation increases not only the operating point, but also the overall neural correlation level,
as indicated by a boost in the intercept while maintaining the slope of the linear relationship in the orthogonal case~\cite{SM}.

In the large-width (mean-field) limit, key parameters $\upgamma_1$, $\upgamma_2$, $\mathcal{K}_1^l$, and $\mathcal{K}_2^l$ can be iteratively constructed from
their values at the input layer. This iteration determines the mean-field solution of the dimension reduction, which captures typical properties of the system under different 
realizations of the network parameters (quenched disorder).
Technical details to derive the above dimensionality evolution for both binary- and continuous-weight cases are given in the supplemental material~\cite{SM}.

\begin{figure}
     \includegraphics[bb=4 4 668 629,width=0.5\textwidth]{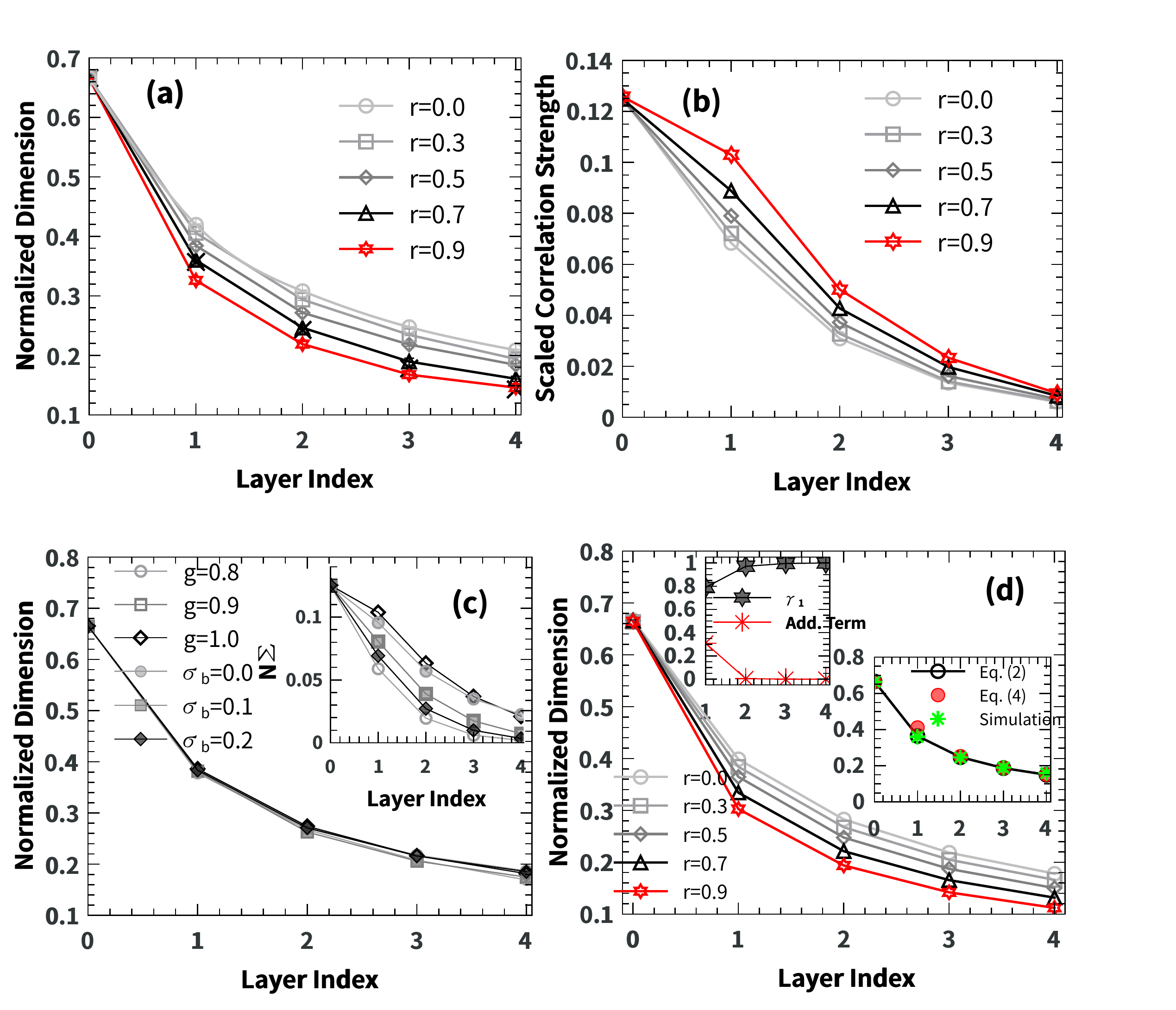}
  \caption{(Color online) Typical behavior of dimension reduction in networks of binary weights. Simulations were carried out on networks of finite size $N=200$, and averaged over
  ten instances with negligible error bars.
  (a) Layer-wise dimension reduction with different correlation level $r$. $g=0.9$, $\alpha=2$, and $\sigma_b=0.1$. The covariance is obtained by Eq.~(\ref{dMFT}).
  The cross symbol indicates the simulation result obtained by layer-wise propagating $10^5$ samples.
  (b) Layer-wise decorrelation with $r$. Other parameters are the same as in (a). The neural correlation strength has been scaled by $N$. (c) Dimension reduction and decorrelation with different values of
  $g$ and $\sigma_b$. $r=0.5$. $g=0.9$ when $\sigma_b$ varies, and $\sigma_b=0.1$ when $g$ varies. (d) Large-$N$ limit behavior for $g=0.4$. The left inset shows the behavior
  of $\upgamma_1$ and the additive term. The right inset shows a comparison of estimated dimensions between theory and simulation ($N=200$). In both insets, $r=0.5$, $\sigma_b=0.1$ and $\alpha=2$. 
  }\label{DR-binary}
\end{figure}

\textit{Results and Discussion.}---
We first study the typical behavior of dimension reduction in networks of binary weights. Surprisingly, the weak correlation among synapses is able to
reduce further the hidden-representation dimensionality across layers compared to the case of orthogonal weights [Fig.~\ref{DR-binary} (a)].
Moreover, the synaptic correlation $r$ can also boost the correlation strength $\Upsigma$ [Fig.~\ref{DR-binary} (b)]. The boost is larger at earlier layers of deep networks. In a practical learning process,
the synaptic plasticity can introduce a certain level of correlations among synapses, reflecting sensory uncertainty. Our theoretical model, despite using a random ensemble of correlated weights,
reveals quantitatively the computation role of the synaptic correlations, which can impact both representation manifold and neural decorrelation process. In other words, 
the weak synaptic correlation accelerates the dimension reduction, while reducing the decay speed of the neural correlation strength.

The exact mechanism underlying the computational roles of synaptic correlation can be revealed by a
large-$N$ expansion of the two-point correlation function. 
In the thermodynamic limit, our theory predicts that the dimensionality can be tuned by two factors: one is multiplicative, captured by $\upgamma_1$, which is observed to grow until arriving at the unity [Fig.~\ref{DR-binary}(d)], and always equal to the unity only at $q=0$.
This multiplicative factor [see Eq.~(\ref{bdim}) or Eq.~(\ref{cdim})] gives rise to further reduction of dimensionality, especially at deeper layers. However, its value can be less than one at earlier layers, 
thereby allowing possibility of increasing the dimensionality at the ($l+1$)-th layer, provided that $N\Upsigma^{l}>\frac{\mathcal{A}}{1-\upgamma_1}-\mathcal{K}_2^{l}$ where $\mathcal{A}>0$ denotes the additive term in Eq.~(\ref{bdim}) or Eq.~(\ref{cdim}).
This predicts that \textit{transient dimensionality expansion is possible} given strong neural correlations or redundant coding at earlier layers, which was empirically observed during training in both feed-forward and recurrent neural networks~\cite{Huang-2018,Eric-2019}.
The other additive factor is directly related to $r^2$. This extra term is clearly positive, thereby contributing an 
additional reduction of dimensionality. Both factors compete with each other; the multiplicative factor saturates at the unity [the left inset of Fig.~\ref{DR-binary} (d)], while the additive term decreases with the network depth, 
which overall makes the dimension reduction
more slowly at deeper networks, in consistent with a practical learning of image classification tasks where the final low-dimensional manifold must be stable for reading out object identities~\cite{Hung-2005,Poggio-2007,Rust-2013,DiCarlo-2016,ID-2019a,ID-2019b}.
\begin{figure}
     \includegraphics[bb=4 4 380 303,scale=0.5]{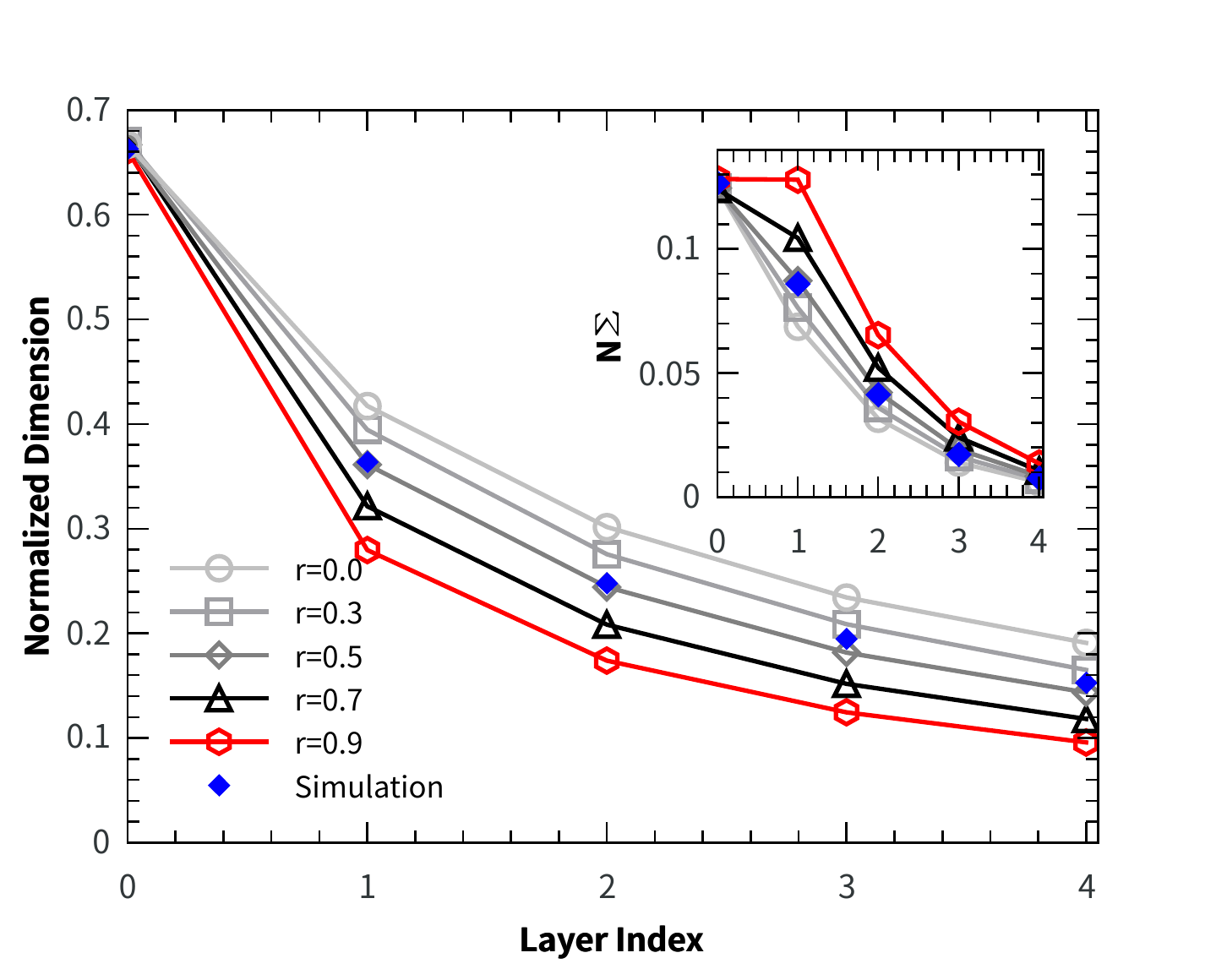}
  \caption{(Color online) Typical behavior of dimension reduction in networks of continuous weights using propagation equation Eq.~(\ref{dMFT}). The results are averaged over ten networks of finite size $N=200$.
  The error bars are negligible. Simulations indicate layer-wise propagating $10^5$ samples.
  $g=0.9$, $\sigma_b=0.1$ and $\alpha=2$.
  }\label{DR-cont}
\end{figure}

The elegant way in which the overall neural correlation level is tuned can also be explained by our theory. The weak synaptic correlation contributes an additional additive term in the linear relationship between the input
and output neural correlations. This additive term, namely $r^2(\mathcal{K}_1^l)^2$ [the same for both binary and continuous weights, see Eq.~(\ref{bsgm}) or Eq.~(\ref{csgm})], increases
the intercept of the linear relationship while maintaining the same slope with the case of $r=0$ (Fig. S1 in~\cite{SM}).
The decorrelation process would thus proceed more slowly when going deeper into the network.
The prediction of our theory may then explain the empirical success in improving the classification performance by a deccorrelation regularization of weights~\cite{ICLR-2017}.

By changing the other statistics of network parameters, say $g$ and $\sigma_b$, but fixing $r$, one can observe rich effects of these parameters [Fig.~\ref{DR-binary} (c)]. First, the dimension reduction seems to be unaffected, or changes by a negligible margin.
However, the correlation strength of the hidden representation displays rich behavior. The weight strength elevates the correlation level, as also expected from previous studies~\cite{Huang-2018}, playing the similar role to the synaptic correlation [Fig.~\ref{DR-binary} (b)].
In contrast, increasing the firing bias would further decorrelate the hidden representation. The mechanisms are encoded into Eq.~(\ref{bsgm}) and Eq.~(\ref{csgm}), in that both the weight strength and firing bias affect the slope and intercept of the linear relationship in a highly non-trivial 
way, via a recursive iteration from initial values of network activity statistics~\cite{SM}. These rich effects of tuning the neural correlation of hidden representations may thus shed light on understanding the empirical success
of reducing overfitting by optimizing decorrelated representations~\cite{ICLR-2015}. The decorrelated representation also coincides with the efficient coding hypothesis in system neuroscience~\cite{Barlow-1961}. In this hypothesis,
a useful representation should be maximally disentangled, rather than being highly redundant. Our analysis provides a theoretical support for this hypothesis, emphasizing the role of weakly-correlated synapses. 

We finally remark that the above analysis also carries over to networks with continuous weights (Fig.~\ref{DR-cont}). Our simulations on deep networks trained with 
Hebbian learning rules~\cite{MacKay-1994,Seung-2017} also show key features of the theoretical predictions of our simple model. Details are given in the 
supplemental material~\cite{SM}.

\textit{Summary.}---Benefits of weakly-correlated synapses are widely claimed in both system neuroscience~\cite{Harris-2013,Harris-2015,Brunel-2016,CorreC-2009} and machine learning~\cite{ICLR-2017}.
The benefits are also realized in a theoretical model of shallow-network unsupervised learning~\cite{Huang-2019a}, a fundamental process in cerebral cortex~\cite{Marr-1970}. However,
the theoretical basis about how weakly-correlated synapses promote the computation in deep neural networks remains unexplored, 
due to the theoretical complexity of handling
the covariance matrix of neural responses. This work tackles the challenge, by deriving mean-field theory for diagonally block-organized 
correlation matrix of synapses, inspired by our recent work of unsupervised learning~\cite{Huang-2019a}.

Our work first identifies the unique scaling form of synaptic correlation level, for both binary and continuous weight values, 
based \textit{solely} on the mathematical self-consistency. This scaling form may be experimentally testable, like in the intact brain~\cite{Bar-2016}.
The theory then reveals in what exact way the weak synaptic correlation tunes both representation dimensionality and associated decorrelation process. More precisely, 
the synaptic correlation accelerates the dimension reduction, while slowing down the 
decorrelation process. Both computation roles can be explained in a large-$N$ expansion of our mean-field equations. 
These predictions coincide with empirical successes in decorrelation regularizations of either synaptic level~\cite{ICLR-2017} or neural level~\cite{ICLR-2015}.
Our theory is thus promising to encourage further understanding of how synaptic and neural correlations interact with each other to yield a disentangled representation supporting the
success of deep learning, and even hierarchical information 
processing in different pathways of neural circuits, a challenging problem in interdisciplinary fields across physics, machine learning and neuroscience~\cite{Saxe-2020,Neuron-2017}.


\begin{acknowledgments}
We thank Chan Li and Wenxuan Zou for a careful reading of the manuscript. This research was supported by the start-up budget 74130-18831109 of the 100-talent-
program of Sun Yat-sen University, and the NSFC (Grant No. 11805284). 
\end{acknowledgments}
\setcounter{figure}{0}    
\renewcommand{\thefigure}{S\arabic{figure}}
\renewcommand\theequation{S\arabic{equation}}
\setcounter{equation}{0}  

\newpage
\onecolumngrid
\appendix

\section*{Supplemental Material}


\section{Theoretical analysis in the large-$N$ limit}
\label{SM-a}

In this paper, we consider deep neural networks of either binary or continuous weights. In the case of binary weights, the distribution of $w_{ij}^l$ is parametrized by zero mean and the covariance given by
\begin{equation}
\overline{w^l_{i j} w^l_{k s}}=\delta_{j s} {q}+\delta_{i k} \delta_{j s}\left(1-{q}\right)={q} \delta_{j s}\left(1-\delta_{i k}\right)+ \delta_{i k} \delta_{j s}\,
\end{equation}
where $\overline{\ \cdot\ }$ denotes the disorder average over the network parameter statistics, and $\delta_{js}$ is the Kronecker delta function. An illustration of this weight covariance is already given in Fig.1 in the main text.
For simplicity, we consider only inter-RF correlations, neglecting intra-RF ones.
The bias parameter $b_i^l$ follows independently a Gaussian distribution with zero mean and variance $\sigma_b$. 
The activation $h_i^l$ is then given by
\begin{equation}
h^{l}_i=\phi\left( \frac{g}{\sqrt{N}}\sum_j w_{ij}^{l}h^{l-1}_j+b^l_i \right)\ .
\end{equation} 
For the continuous-weight case, the parameter $w_{ij}^l$ follows the Gaussian distribution with zero mean and the covariance specified by
\begin{equation}
\overline{w^l_{i j} w^l_{k s}}=\delta_{j s} {q}+\delta_{i k} \delta_{j s}\left(\frac{g^2}{N}-{q}\right)={q} \delta_{j s}\left(1-\delta_{i k}\right)+\frac{g^2}{N} \delta_{i k} \delta_{j s} .
\end{equation}
The bias parameter $b_i^l$ follows independently a Gaussian distribution with zero mean and variance $\sigma_b$. The activation $h_i^l$ is then given by
\begin{equation}
h^{l}_i=\phi\left( \sum_j w_{ij}^{l}h^{l-1}_j+b^l_i \right)\ .
\end{equation}


\subsection{Mean-field iteration of activity moments}
\label{SM-a1}
In this section, we derive the mean-field iteration of activity moments. Note that, for both binary and continuous weights, the form of the iteration looks very similar, 
we here thus derive in detail the iteration for the continuous case. 
We first derive the mean-field equation for the mean activity $m_i^l$ as follows,
\begin{equation}
\begin{aligned} m_i^l &= \langle h_i^l\rangle\\ &= \left\langle \phi \left( \left[\mathbf{w}^{l} \mathbf{h}^{l-1}\right]_{i} +b_i^l \right) \right\rangle  \\&=  \left\langle \phi \left( a_i^l +\left[\mathbf{w}^{l} \mathbf{m}^{l-1}\right]_{i} +b_i^l \right) \right\rangle \ , \end{aligned}
\end{equation}
where the average $\left<\cdot\right>$ is defined over the activity statistics throughout the paper, and we define the mean-subtracted weighted-sum (or pre-activation) $a_{i}^{l}=\sum_{j} w_{i j}^{l}\left(h_{j}^{l-1}-\left\langle h_{j}^{l-1}\right\rangle\right)$, then its expectation is zero, and variance is given by
$ \Delta_{i j}^{l}=\left\langle a_{i}^{l} a_{j}^{l}\right\rangle=\left[\mathbf{w}^{l} \mathbf{C}^{l-1}\left(\mathbf{w}^{l}\right)^{T}\right]_{i j}$, where $\mathbf{C}$ denotes the covariance matrix of the neural activity. Because $a_i^l$ is the sum of $N$ nearly-independent random terms, as $N\to \infty$, we apply the central limit theorem, and obtain
\begin{equation}
\begin{aligned} 
m_i^l &=  \int Dt\phi \left( \sqrt{\Delta_{ii}^l}t  +\sum_j w_{ij}^l m_j^{l-1} +b_i^l\right) \ ,   
\end{aligned}\label{mean activity}
\end{equation}
where $Dt = e^{-t^2/2}dt/\sqrt{2\pi}$.
Then, we consider the covariance of activities. Note that the Gaussian random variable $a_i^l$ has a variance $\Delta_{ii}^l$. The activity covariance is then given by
\begin{equation}
\begin{aligned}  
C_{ij}^l  &=\left\langle h_{i}^{l} h_{j}^{l}\right\rangle-\left\langle h_{i}^{l}\right\rangle\left\langle h_{j}^{l}\right\rangle \\
&=\left\langle\phi\left(a_{i}^{l}+\left[\mathbf{w}^{l} \mathbf{m}^{l-1}\right]_{i}+b_{i}^{l}\right) \phi\left(a_{j}^{l}+\left[\mathbf{w}^{l} \mathbf{m}^{l-1}\right]_{j}+b_{j}^{l}\right)\right\rangle- m_{i}^{l} m_{j}^{l}\\ 
&= \int D x D y \phi\left(\sqrt{\Delta_{i i}^{l}} x+b_{i}^{l}+\left[\mathbf{w}^{l} \mathbf{m}^{l-1}\right]_{i}\right) \phi\left(\sqrt{\Delta_{j j}^{l}}(\uppsi x+y \sqrt{1-\uppsi^{2}} ) \right.
\\&\quad\left. +b_{j}^{l}+\left[\mathbf{w}^{l} \mathbf{m}^{l-1}\right]_{j} \right)  -m_{i}^{l} m_{j}^{l} \ ,
\end{aligned}\label{covariance of activities}
\end{equation}
where $D x={e^{-x^{2} / 2} d x }/{ \sqrt{2 \pi}}$, and $\uppsi={\Delta_{i j}^{l}}/{\sqrt{\Delta_{i i}^{l} \Delta_{j j}^{l}}}$. $a_i^l$ and $a_j^l$ have been parametrized by two independent standard Gaussian random variables, say $x$ and $y$, respectively. The pre-activation 
correlation has been captured by the correlation coefficient $\uppsi$ ($|\uppsi|\le 1$).

For binary weights, the above Eqs.~(\ref{mean activity}), (\ref{covariance of activities}) become
\begin{equation}
\begin{aligned} 
m_i^l &=  \int Dt\phi \left( \sqrt{\Delta_{ii}^l}t  +\frac{g}{\sqrt{N}}\sum_j w_{ij}^l m_j^{l-1} +b_i^l\right) \ ,   
\end{aligned}\label{mean activity binary}
\end{equation}
and
\begin{equation}
\begin{aligned}  
C_{ij}^l  &= \int D x D y \phi\left(\sqrt{\Delta_{i i}^{l}} x+b_{i}^{l}+ \frac{g}{\sqrt{N}} \left[\mathbf{w}^{l} \mathbf{m}^{l-1}\right]_{i}\right) \phi\left(\sqrt{\Delta_{j j}^{l}}(\uppsi x+y \sqrt{1-\uppsi^{2}} ) \right.
\\&\quad\left. +b_{j}^{l}+   \frac{g}{\sqrt{N}} \left[\mathbf{w}^{l} \mathbf{m}^{l 1}\right]_{j} \right)  -m_{i}^{l} m_{j}^{l} \ ,
\end{aligned}\label{covariance of activities binary}
\end{equation}
where 
\begin{equation}
  \Delta_{i j}^{l}=\frac{g^2}{N}\left[\mathbf{w}^{l} \mathbf{C}^{l-1}\left(\mathbf{w}^{l}\right)^{T}\right]_{i j}.
\end{equation}

With the activity moments, we can then evaluate the dimensionality of the $l$-th layer by

\begin{equation}
D^l=\frac{\left ( \sum_i \lambda_i \right)^2 }{\sum_i \lambda_i^2}= \frac{\left(\operatorname{Tr}\mathbf{C}^l\right)^2}{\operatorname{Tr}(\mathbf{C}^l)^2}=\frac{\left( \sum_i C_{ii}^l\right)^2 }{\sum_{i,j}(C_{ij}^l)^2},
\end{equation}
where $\{\lambda_i\}$ is the eigen-spectrum of the covariance matrix $\mathbf{C}^l$. Then we can define the normalized dimensionality as $\tilde{D}^l=\frac{(\operatorname{Tr}\mathbf{C}^l)^2 }{N \operatorname{Tr}(\mathbf{C}^l)^2}$, which
is then independent of the network width $N$. To derive the recursion of dimensionality for each layer, we define additionally $\mk_1^l =\frac{1}{N}\sum_iC_{ii}^l$, $\mk_2^l = \frac{1}{N}\sum_{i}(C_{ii}^l)^2$, and $\Upsigma^l=\frac{2}{N^2}\sum_{i<j}(C_{ij}^l)^2$ for a large value of $N$.
The normalized dimensionality of the $l$-th layer is
thus expressed as
\begin{equation}
\label{OriD}
\tilde{D} ^l =\frac{(\mk_1^l)^2}{N\Upsigma^l+\mk_2^l},
\end{equation}
which is useful for the following theoretical analysis.

\subsection{Expansion of two-point correlations }
\label{Expan-corr}
In the mean-field limit, we can assume $C_{ij}^l\sim \mathcal{O}(1/\sqrt{N})$ for $i\ne j$~\cite{Mezard-1987}. In the following part, we assume weights take continuous values. Binary weights can be similarly analyzed, by noting that the pre-activations should be multiplied by a pre-factor $g/\sqrt{N}$.

We first analyze the off-diagonal part of the covariance matrix. First, we notice that $\overline{\Delta_{i j}^{2}}=\sum_{k, l}\overline{w_{i k}^{2}}\overline{w_{j l}^{2}} C_{k l}^{2} =N^{2} \frac{g^2}{N} \frac{g^2}{N} \frac{1}{N} \sim \mathcal{O} \left(\frac{g^4}{N}\right)$,
which means that $\Delta_{ij}\sim \mathcal{O}(\frac{g^2}{\sqrt{N}})$. That is, when $N$ is sufficiently large, $\Delta_{ij}$ is very small. Then we can carry out a Taylor expansion with respect to
a small $\Delta_{ij}$ whose layer index is added below:

\begin{equation}
\begin{aligned}
\phi\left(\sqrt{\Delta_{j j}^{l}}(\uppsi x+y \sqrt{1-\uppsi^{2}})+z_j^0\right) &=\phi\left(\sqrt{\Delta_{j j}^{l}} y+z_j^0\right) \\
& +\phi^{\prime}\left(\sqrt{\Delta_{j j}^{l}} y+z_j^0\right) \frac{x \Delta_{i j}^{l}}{\sqrt{\Delta_{i i}^{l}}}+\mathcal{O}\left(\left(\Delta_{i j}^{l}\right)^{2}\right)  ,
\end{aligned}
\end{equation}
where we define $z_j^0=b_{j}^{l}+\left[\mathbf{w}^{l} \mathbf{m}^{l-1}\right]_{j}$. By noting that $ m_i^l =  \int Dt\phi \left( \sqrt{\Delta_{ii}^l}t + z_i^0\right) $, we obtain

\begin{equation}
\label{K_ijeq}
\begin{aligned} C_{ij}^l&=\int D x D y \phi\left(\sqrt{\Delta_{i i}^{l}} x+z_i^0\right) \phi^{\prime}\left(\sqrt{\Delta_{j j}^{l}} y+z_j^0\right) \frac{x \Delta_{i j}^{l}}{\sqrt{\Delta_{i i}^{l}}}+\mathcal{O}\left(\left(\Delta_{i j}^{l}\right)^{2}\right)\\
&=\int D x D y \phi'\left(\sqrt{\Delta_{i i}^{l}} x+z_i^0\right) \phi'\left(\sqrt{\Delta^l_{jj}}y+z_j^0\right)\Delta_{ij}^l+\mathcal{O}\left(\left(\Delta_{ij}^l\right)^2\right) . \end{aligned}
\end{equation}
Therefore we can write $C_{ij}^l\simeq\left<\phi'\left(\sqrt{\Delta_{i i}^{l}} x+z_i^0\right) \right>_x\left<\phi'\left(\sqrt{\Delta_{jj}^{l}} x+z_j^0\right)\right>_y\Delta_{ij}^l$, where the linear coefficient is an average over standard normal variables, and is called hereafter $K_{ij}^l$ for the following analysis.

We next remark that $\overline{\Delta_{ii}}\simeq\frac{g^2}{N}\sum_k\overline{w_{ik}^2}C_{kk}=g^2\mk_1\sim\mathcal{O}(g^2)$. In the small-$g$ limit,
we can carry out an expansion in $\sqrt{\Delta_{ii}}$ whose layer index is added below, and get
\begin{equation}
\begin{aligned} 
C_{ij}^l&\simeq \int D x D y \left[\phi'\left(z_i^0\right) +\phi''  \left(z_i^0\right) \sqrt{\Delta_{i i}^{l}} x\right]\left[ \phi'\left(z_j^0\right)+ \phi''\left(z_j^0\right) \sqrt{\Delta^l_{jj}}y \right]\Delta_{ij}^l \\
&= \phi'(z_i^0) \phi'(z_j^0) \Delta_{ij}^l .
\end{aligned}\label{covariance expanded}
\end{equation}
This result was reported in our previous work~\cite{Huang-2018}. We then analyze the diagonal part of the covariance matrix,
\begin{equation}
\begin{aligned}  
C_{ii}^l  &=\left\langle h_{i}^{l} h_{i}^{l}\right\rangle-\left\langle h_{i}^{l}\right\rangle\left\langle h_{i}^{l}\right\rangle \\ &=\left\langle\phi^2\left(a_{i}^{l}+z_i^0\right) \right\rangle- m_{i}^{l} m_{i}^{l}
\\ &= \int D x  \phi^2 \left(\sqrt{\Delta_{i i}^{l}} x+z_i^0\right) -
\int D x  \phi \left(\sqrt{\Delta_{i i}^{l}} x+z_i^0\right)
\int D y  \phi \left(\sqrt{\Delta_{i i}^{l}} y +z_i^0\right) .
\end{aligned}
\end{equation}
We expand the above formula in the small $\Delta_{ii}^l$, i.e., $\phi\left(a_{i}^{l}+z_i^0\right)  = \phi\left(z_i^0\right) +\phi'  \left(z_i^0\right) \sqrt{\Delta_{i i}^{l}} x$, and obtain

\begin{equation}\label{covexp2}
\begin{aligned}C_{ii}^l &\simeq  \int Dx\left[   \phi\left(z_i^0\right) +\phi'  \left(z_i^0\right) \sqrt{\Delta_{i i}^{l}} x \right]^2-\left[\int Dx\left( \phi\left(z_i^0\right) +\phi'  \left(z_i^0\right) \sqrt{\Delta_{i i}^{l}} x\right)\right]^2\\
&=  \left[\phi'( z_i^0)\right]^2\Delta_{ii}^l . 
\end{aligned}
\end{equation}
Therefore, we can write $C_{ii}^l\simeq K_{ii}^l\Delta_{ii}^l$, where $K_{ii}^l$ is the shorthand for the linear coefficient. To improve the prediction accuracy, one needs to include high-order terms into this approximation.
We observe that if we use Eq.~(\ref{K_ijeq}) by setting $i=j$, the theoretical prediction in the main text can match the numerical simulation results even in a relatively large value of $g$. This may be due to the fact that
the contribution of $\Delta_{ii}^l$ is taken into account when computing $K_{ii}^l$.
For binary weights, the expanded covariance is just the same as the equations [Eq.~(\ref{K_ijeq}) and Eq.~(\ref{covexp2})], yet with $z_i^0=b_{i}^{l}+\frac{g}{\sqrt{N}}\left[\mathbf{w}^{l} \mathbf{m}^{l-1}\right]_{i}$.

\subsection{Iteration of the correlation strength $\Upsigma^l$}
\subsubsection{Binary weights}
\label{SM_bin}
First, we calculate $\mk_1^l$, and in the large $N$ and small $g$ limits, we obtain
\begin{equation}
\begin{aligned}\mk_1^l &=\overline{\left(\phi^{\prime}\left(\frac{g}{\sqrt{N}}\sum_{j=1}^{N} w_{i j}^{l} m_{j}^{l-1}+b_{i}^{l}\right)\right)^{2} }\Delta_{i i}^l\\
& \simeq  \overline{\left(\phi^{\prime}\left( \frac{g}{\sqrt{N}} \sum_{j=1}^{N} w_{i j}^{l} m_{j}^{l-1}+b_{i}^{l}\right)\right)^{2} } g^2 \mk_1^{l-1}, \end{aligned}
\end{equation}
where $\overline{\ \cdot\ }$ means an average over the distribution of network parameters, and $\Delta_{ii}^l$ is approximated by 
\begin{equation}
\Delta_{ii}^l\simeq\frac{g^2}{{N}}\overline{\sum_{k, j}^{N} w_{i k}^l w_{i j}^l C_{k j}^{l-1}}=\frac{g^2}{N} \sum_{k=1}^{N} C_{kk}^{l-1}=g^2\mk_1^{l-1}\ .
\end{equation}
Note that the argument of $\phi'(\cdot)$ is a sum of a large number of nearly-independent random variables. It is then easy to write that $\overline{ \frac{g}{\sqrt{N}}\sum_j w_{ij}^l m_j^{l-1}+b_i^l} = 0$, 
and $$\overline{  \left( \frac{g}{\sqrt{N}} \sum_j w_{ij}^l m_j^{l-1}+b_i^l\right)^2} = \frac{g^2}{N}\sum_j \left(m_j^{l-1}\right)^2 +\sigma_b \ .$$ 
According to the central limit theorem, we obtain
\begin{equation}
\overline{ \left(\phi^{\prime}\left( \frac{g}{\sqrt{N}} \sum_{j=1}^{N} w_{i j}^{l} m_{j}^{l-1}+b_{i}^{l}\right)\right)^{2}}=\int D x\left(\phi^{\prime}(x \sqrt{g^2 Q^{l-1}+\sigma_{b}})\right)^{2}\defe \overline{ K_{ii}^{l}}\ ,
\end{equation}
where we have defined $Q^{l-1}\defe \frac{1}{N}\sum_{i=1}^N \left(m_i^{l-1}\right)^2$.
The recursion of $Q^l$ becomes
\begin{equation}
\begin{aligned}Q^l &=  \frac{1}{N}\sum_i\left[ \int Dt\phi \left( \sqrt{\Delta_{ii}^l}t  + \frac{g}{\sqrt{N}} \sum_j w_{ij} m_j^{l-1} +b_i^l\right)\right]^2\\
 &=\int Dx \left[   \int Dt \phi \Biggl(\sqrt{g^2\mk_1^{l-1}}t+x\sqrt{g^2 Q^{l-1}+\sigma_b}\Biggr)\right ]^2,
\end{aligned}
\end{equation}
where we have used $\Delta_{ii}^l = g^2\mk_1^{l-1}$.

Finally, we obtain the recursion for $\mk_1^l$,
\begin{equation}
\mk_1^l = g^2 \overline{K_{ii}^l} \mk_1^{l-1}. \label{K1 binary}
\end{equation}
Note that $\mk_1^l$ can also be calculated recursively without the small-$g$ assumption as follows,
\begin{equation}
\begin{aligned}
\mk_1^l&=\int Dx Dt\phi^2\left(\sqrt{g^2\mk_1^{l-1}}t+x\sqrt{\sigma_b +g^2 Q^{l-1}}\right)-Q^l\\
&=\int Dx \phi^2\left( \sqrt{g^2\mk_1^{l-1}+\sigma_b +g^2 Q^{l-1}}\ x \right)-Q^l.
\end{aligned}
\end{equation}

Next, the recursion of $\mk_2^l$ can be calculated by definition as follows,
\begin{equation}
\begin{aligned}
\mk_2^l &= \frac{1}{N}\sum_{i}\left(C_{ii}^l\right)^2
\\ &= \frac{1}{N}\sum_{i}\left(K_{ii}^l g^2 \mk_1^{l-1}\right)^2
\\ & = \overline{(K_{ii}^l)^2} g^4 (\mk_1^{l-1})^2 \ ,
\end{aligned} \label{K2 binary}
\end{equation}
where we have assumed that $K_{ii}^l$ in the large-$N$ limit does not depend on the specific site index, and thus
\begin{equation}
\overline{(K_{ii}^l)^2} = \overline{\left(\phi^{\prime}\left(\frac{g}{\sqrt{N}}\sum_{j} w_{i j}^{l} m_{j}^{l-1}+b_{i}^{l}\right)\right)^{4}}=\int D x\left(\phi^{\prime}(x \sqrt{g^2 Q^{l-1}+\sigma_{b}})\right)^{4} \ .
\end{equation}
Note that $\mk_2^l$ can be evaluated recursively without the small-$g$ assumption as
\begin{equation}
 \mk_2^{l}=\left<\left[\left<\phi^2(f)\right>_z-\left<\phi(f)\right>_z^2\right]^2\right>_{u,t},
\end{equation}
where $z$, $u$, and $t$ are all standard normal variables, and $f\defe\sqrt{g^2\mk_1^{l-1}}z+\sqrt{\sigma_b}u+\sqrt{g^2Q^{l-1}}t$.

We finally derive the recursion of $\Upsigma^l$. First, for the binary weights, to compute $\Delta_{ij}^2$, where the layer index can be added later,
we have
\begin{equation}
\begin{aligned}
\left(\sum_{k,l}w_{ik}w_{jl}C_{kl}\right)^2 &= \left(\sum_{k\neq l}w_{ik}w_{jl}C_{kl}+\sum_{k}w_{ik}w_{jk}C_{kk}\right)^2\\
&\simeq \sum_{k\neq l;k'\neq l'}w_{ik}w_{ik'}w_{jl}w_{jl'}C_{kl}C_{k'l'}+\sum_{k,k'}w_{ik}w_{jk}w_{ik'}w_{jk'}C_{kk}C_{k'k'}\\
&\simeq (1+q^2)\sum_{k\neq l}C_{kl}^2+(1-q^2)\sum_{k}C_{kk}^2+q^2\left(\sum_kC_{kk}\right)^2,
\end{aligned}
\end{equation}
where the cross-term vanishes in statistics to derive the second equality, due to the vanishing intra-RF correlation for one hidden neuron. The third equality is derived by considering the inter-RF correlation in our current setting.
Finally, we arrive at
\begin{equation}
\begin{aligned}N\Upsigma^{l+1} &=\frac{2}{N}\sum_{i<j}\overline{(K_{ij}^{l+1})^2}\frac{g^4}{N^2}\left[(1+q^2)N^2\Upsigma^l+(1-q^2)N\mk_2^l+q^2N^2\left(\mk_1^l\right)^2\right]\\
&=\overline{(K_{ij}^{l+1})^2}g^4 \left[  (1+q^2)N\Upsigma^{l}+q^2 N(\mk_1^{l})^2+(1-q^2)\mk_2^{l}  \right]\\
&\simeq  \overline{(K_{ij}^{l+1})^2} g^4\left[  N\Upsigma^l+\mk_2^l + r^2 (\mk_1^l)^2\right]\ ,
\end{aligned} \label{Sigma binary}
\end{equation}
where we have to assume $q=r/\sqrt{N}$ [$r\sim\mathcal{O}(1)$], and $\overline{(K_{ij}^{l+1})^2}$ is used to replace $(K_{ij}^{l+1})^2$ in the mean-field approximation
and can be computed recursively as follows,
\begin{equation}
\begin{aligned}
\overline{(K_{ij}^{l+1})^2} &= \left< \left<\phi'\left(\sqrt{g^2\mk_1^l} x+\sqrt{g^2Q^l}z_1+\sqrt{\sigma_b}u_1\right) \right>^2_x\right.\\
&\left.\times\left<\phi'\left(\sqrt{g^2\mk_1^l} y+\sqrt{g^2Q^l}\left(\rho z_1+\sqrt{1-\rho^2}z_2\right)+\sqrt{\sigma_b}u_2\right)\right>^2_y\right> _{z_1,z_2,u_1,u_2},
\end{aligned}
\end{equation}
where $x,y,z_1,z_2,u_1,u_2$ are all standard Gaussian random variables, capturing both thermal and disorder average (inner and outer ones respectively). The correlation coefficient is given by
\begin{equation}
 \rho\defe\frac{\overline{(z_i^0-b_i)(z_j^0-b_j)}}{\sqrt{\overline{(z_i^0-b_i)^2}\cdot\overline{(z_j^0-b_j)^2}}}=q.
\end{equation}
Therefore, $q$ plays an important role in our current theory, in contrast to the orthogonal case~\cite{Huang-2018}.

\subsubsection{Continuous weights}
\label{SM_cont}

First, we calculate $\mk_1^l$, and in the large $N$ limit, we obtain

\begin{equation}
\begin{aligned}
\mk_1^l &=
\overline{\left(\phi^{\prime}\left(\sum_{j} w_{i j}^{l} m_{j}^{l-1}+b_{i}^{l}\right)\right)^{2} }\Delta_{i i}^l
\\& \simeq \overline{\left(\phi^{\prime}\left(\sum_{j} w_{i j}^{l} m_{j}^{l-1}+b_{i}^{l}\right)\right)^{2} } g^2 \mk_1^{l-1},
\end{aligned}
\end{equation}
where $\overline{\ \cdot\ }$ means an average over the distribution of network parameters, and $\Delta_{ii}^l$ can be approximated by
\begin{equation}
\Delta_{ii}^l\simeq\overline{\sum_{k, j}^{N} w_{i k}^l w_{i j}^l C_{k j}^{l-1}}=\frac{g^2}{N} \sum_{k=1}^{N} C_{kk}^{l-1}=g^2\mk_1^{l-1}\ .
\end{equation}
The argument of $\phi'(\cdot)$ is a sum of a large number of nearly-independent random variables, as a result, it is easy to write that $\overline{ \sum_j w_{ij}^l m_j^{l-1}+b_i^l} = 0$, and
$$\overline{  \left( \sum_j w_{ij}^l m_j^{l-1}+b_i^l\right)^2} = \frac{g}{N}\sum_j \left(m_j^{l-1}\right)^2 +\sigma_b \ .$$ 
According to the central limit theorem, we obtain
\begin{equation}
\overline{ \left(\phi^{\prime}\left(\sum_{j=1}^{N} w_{i j}^{l} m_{j}^{l-1}+b_{i}^{l}\right)\right)^{2}}=\int D x\left(\phi^{\prime}(x \sqrt{g^2 Q^{l-1}+\sigma_{b}})\right)^{2}\defe \overline{ K_{ii}^{l}}\ ,
\end{equation}
where we have defined $Q^{l-1}\defe \frac{1}{N}\sum_i \left(m_i^{l-1}\right)^2$.
The recursion of $Q^l$ becomes
\begin{equation}
\begin{aligned}
Q^l &=  \frac{1}{N}\sum_i\left[ \int Dt\phi \left( \sqrt{\Delta_{ii}^l}t  +\sum_j w_{ij} m_j^{l-1} +b_i^l\right)\right]^2\\
&=\int Dx \left[   \int Dt \phi \Biggl(\sqrt{g^2 \mk_1^{l-1}}t+x\sqrt{g^2 Q^{l-1}+\sigma_b}\Biggr)\right ]^2,
\end{aligned}
\end{equation}
where we have used $\Delta_{ii}^l = g^2\mk_1^{l-1}$.

Finally, we obtain the recursion for $\mk_1^l$ as

\begin{equation}
\mk_1^l = g^2 \overline{K_{ii}^l} \mk_1^{l-1} \label{K1 continuous}.
\end{equation}
Note that $\mk_1^l$ can also be calculated recursively without the small-$g$ assumption as follows,
\begin{equation}
\mk_1^l=\int Dx \phi^2\left( \sqrt{g^2\mk_1^{l-1}+\sigma_b +g^2 Q^{l-1}}\ x \right)-Q^l.
\end{equation}

We then derive $\mk_2^l$ as follows,
\begin{equation}
\begin{aligned}
\mk_2^l &= \frac{1}{N}\sum_{i}\left(C_{ii}^l\right)^2
\\ &= \frac{1}{N}\sum_{i}\left(K_{ii}^l g^2 \mk_1^{l-1}\right)^2
\\ & = \overline{(K_{ii}^l)^2} g^4 (\mk_1^{l-1})^2 \ ,
\end{aligned}  \label{K2 continuous}
\end{equation}
where $(K_{ii}^l)^2$ is approximated by
\begin{equation}
\overline{(K_{ii}^l)^2} = \overline{\left(\phi^{\prime}\left(\sum_{j} w_{i j}^{l} m_{j}^{l-1}+b_{i}^{l}\right)\right)^{4}}=\int D x\left(\phi^{\prime}(x \sqrt{g^2 Q^{l-1}+\sigma_{b}})\right)^{4} \ .
\end{equation}
Note that $\mk_2^l$ can be evaluated recursively without the small-$g$ assumption as
\begin{equation}
 \mk_2^{l}=\left<\left[\left<\phi^2(f)\right>_z-\left<\phi(f)\right>_z^2\right]^2\right>_{u,t},
\end{equation}
where $z$, $u$, and $t$ are all standard normal variables, and $f\defe\sqrt{g^2\mk_1^{l-1}}z+\sqrt{\sigma_b}u+\sqrt{g^2Q^{l-1}}t$.

To finally obtain the recursion of $\Upsigma^l$, we first analyze $\Delta_{ij}^2$ whose layer index will be added later as follows,
\begin{equation}
\begin{aligned}
\Delta_{ij}^2&=\left(\sum_{k,l}w_{ik}w_{jl}C_{kl}\right)^2 = \left(\sum_{k\neq l}w_{ik}w_{jl}C_{kl}+\sum_{k}w_{ik}w_{jk}C_{kk}\right)^2\\
&\simeq \sum_{k\neq l;k'\neq l'}w_{ik}w_{ik'}w_{jl}w_{jl'}C_{kl}C_{k'l'}+\sum_{k,k'}w_{ik}w_{jk}w_{ik'}w_{jk'}C_{kk}C_{k'k'}\\
&\simeq\left(\frac{g^4}{N^2}+q^2\right)\sum_{k\neq l}C_{kl}^2+\frac{g^4}{N^2}(2\rho^2+1)\sum_kC_{kk}^2\\
&+q^2\left[\left(\sum_kC_{kk}\right)^2-\sum_kC_{kk}^2\right],
\end{aligned}   
\end{equation}
where $\rho=\frac{qN}{g^2}$ due to the fact that $\overline{w_{ik}w_{jk}}=q$ as well as $\overline{w_{ik}^2}=g^2/N$. To arrive at the final equality, we have used the statistics
equality as follows,
\begin{equation}
\begin{aligned}
 \overline{w_{ik}^2w_{jk}^2}&=\left<\frac{g^2}{N}z_1^2\times\frac{g^2}{N}(\rho z_1+\sqrt{1-\rho^2}z_2)^2\right>_{z_1,z_2\sim\mathcal{N}(0,1)}\\
 &=\frac{g^4}{N^2}(2\rho^2+1).
 \end{aligned}
\end{equation}
We finally arrive at the recursion for $\Upsigma^l$ as follows,
\begin{equation}
\begin{aligned}
N\Upsigma^l &\simeq\frac{2}{N}\sum_{i<j}\overline{(K_{ij}^l)^2}\left[\left(\frac{g^4}{N^2}+q^2\right)\sum_{k\neq j}(C_{kj}^{l-1})^2+\left(q^2+\frac{g^4}{N^2}\right)\sum_k(C_{kk}^{l-1})^2+q^2\left(\sum_kC_{kk}^{l-1}\right)^2\right]\\
&=\overline{(K_{ij}^l)^2} \left[(g^4 +N^2q^2)(N\Upsigma^{l-1}+\mk_2^{l-1})+q^2N^3 (\mk_1^{l-1})^2\right]
\\&= \overline{(K_{ij}^l)^2} \left[ g^4 (N\Upsigma^{l-1} +\mk_2^{l-1})+r^2(\mk_1^{l-1})^2 \right]\ ,
\end{aligned}   \label{Sigma continuous}
\end{equation}
where we have to assume $q=r/N^{\frac{3}{2}}$ [$r\sim\mathcal{O}(1)$], and $\overline{(K_{ij}^l)^2}$ is used to approximate $(K_{ij}^l)^2$ and can be
computed as follows,
\begin{equation}
\begin{aligned}
\overline{(K_{ij}^{l})^2} &= \left< \left<\phi'\left(\sqrt{g^2\mk_1^{l-1}} x+\sqrt{g^2Q^{l-1}}z_1+\sqrt{\sigma_b}u_1\right) \right>^2_x\right.\\
&\left.\times\left<\phi'\left(\sqrt{g^2\mk_1^{l-1}} y+\sqrt{g^2Q^{l-1}}\left(\rho z_1+\sqrt{1-\rho^2}z_2\right)+\sqrt{\sigma_b}u_2\right)\right>^2_y\right> _{z_1,z_2,u_1,u_2},
\end{aligned}
\end{equation}
where $x,y,z_1,z_2,u_1,u_2$ are all standard Gaussian random variables, capturing both thermal and disorder average (inner and outer ones respectively). The correlation coefficient in the continuous-weight case
is given by
\begin{equation}
 \rho\defe\frac{\overline{(z_i^0-b_i)(z_j^0-b_j)}}{\sqrt{\overline{(z_i^0-b_i)^2}\cdot\overline{(z_j^0-b_j)^2}}}=\frac{qN}{g^2}.
\end{equation}
\begin{figure}
	\centering
	\includegraphics[bb=2 7 400 305,scale=0.6]{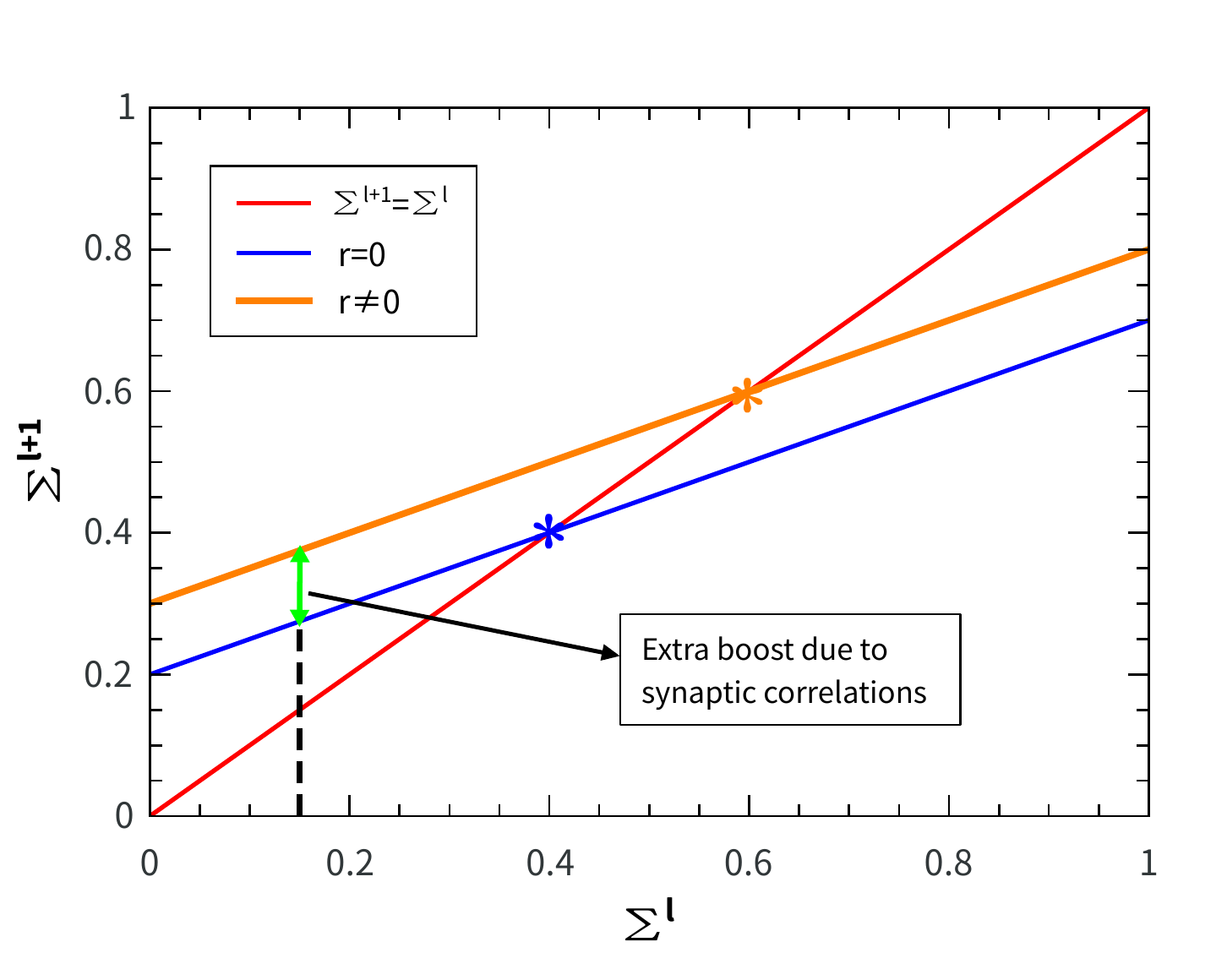}
	\caption{The schematic illustration showing how synaptic correlations elevate the neural correlation level (multiplied by $N$) and the operating point in
	hidden representations of deep neural networks. The boost is indicated by the double arrow for an example in which the input $\Upsigma^l$
	is below the operating point (indicated by star-symbols) where $\Upsigma^{l+1}=\Upsigma^l$.}
	\label{boost}
\end{figure}

We finally remark that the synaptic correlation is able to boost the neural correlation level when transmitting signal via hidden representations.
From the linear relationship between $\Upsigma^{l+1}$ and $\Upsigma^l$ [see Eq.~(\ref{Sigma binary})], one derives for the binary weights
that the operating point is given by:
\begin{equation}
\label{oper}
 \Upsigma_{*}^l=\frac{\Upsilon\mk_2^l}{1-\Upsilon}+\frac{\Upsilon r^2(\mk_1^l)^2}{1-\Upsilon},
\end{equation}
where $\Upsigma_{*}^l$ has been multiplied by $N$, and $\Upsilon\defe g^4\overline{(K_{ij}^{l+1})^2}$. Eq.~(\ref{oper}) implies that the operating point is increased by the synaptic correlations (the last term in the equation).
The intercept of the linear relationship is also increased by
a positive amount $\Upsilon r^2(\mk_1^l)^2$. Note that the slope of the linear relationship under the orthogonal-weight and correlated-weight cases are the same.
These effects are qualitatively the same for both continuous and binary weights, which is shown in Fig.~\ref{boost}.

\subsection{Iteration of the dimensionality across layers}

\subsubsection{Binary weights}

According to the definition, with the help of Eqs.~(\ref{K1 binary}) and~(\ref{K2 binary}) and the recursion equation for $\Upsigma^l$, we obtain

\begin{equation}
\begin{aligned}
\tilde{D}^l&= \frac{\left(\mk_1^l\right)^2}{N\Upsigma^l+\mk_2^l}
\\&= 
\frac{(\mk_1^{l-1})^2}{\upgamma_1(N\Upsigma ^{l-1}+\mk_2^{l-1})+\left(\upgamma _1 {r^2}+\upgamma_2\right)\left (\mk_1^{l-1}\right )^2}\ ,
\end{aligned} \label{Iteration of D binary }
\end{equation}
where $\upgamma_1= \overline{K_{ij}^2}/{\overline{K_{ii}}^2}$, $\upgamma_2= \overline{K_{ii}^2}/\overline{K_{ii}}^2$ and $r=qN^{\frac{1}{2}}$.
When the superscripts of layer index for $K_{ij}$ and $K_{ii}$ are clear, the superscripts are omitted. Here, we manage to use the activity statistics at previous layers to estimate
the dimensionality of the current layer, rather than the original formula [Eq.~(\ref{OriD})]. Thus the mechanism for dimensionality change can be revealed.

Note that to evaluate $\upgamma_1$ and $\upgamma_2$, we need to compute the following quantities,
\begin{equation}\label{g1g2eq}
\begin{aligned}
\overline{K_{ii}}&=\left<\left(\phi^{\prime}\left(x \sqrt{g ^2Q ^{l-1} +\sigma_{b}}\right)\right)^{2}\right>_x\ ,\\
\overline{K_{ii}^2}&= \left<\left(\phi^{\prime}\left(y \sqrt{g^2  Q^{l-1}+\sigma_{b}}\right)\right)^{4}\right>_y \ , \\
\overline{K_{ij}^2}&=\left< \left<\phi'\left(\sqrt{g^2\mk_1^{l-1}} x+\sqrt{g^2Q^{l-1}}z_1+\sqrt{\sigma_b}u_1\right) \right>^2_x\right.\\
&\left.\times\left<\phi'\left(\sqrt{g^2\mk_1^{l-1}} y+\sqrt{g^2Q^{l-1}}\left(\rho z_1+\sqrt{1-\rho^2}z_2\right)+\sqrt{\sigma_b}u_2\right)\right>^2_y\right> _{z_1,z_2,u_1,u_2} \ ,
\end{aligned}
\end{equation}
where $\rho=q$. $Q^l$, $\mk_1^l$, and $\mk_2^l$ can also be computed recursively by following the iterative equations in Sec.~\ref{SM_bin}.

\subsubsection{Continuous weights}

According to the definition, with the help of Eqs.~(\ref{K1 continuous}) and~(\ref{K2 continuous}) and the recursion equation for $\Upsigma^l$, we obtain

\begin{equation}
\begin{aligned}
\tilde{D}^l&= \frac{\left(\mk_1^l\right)^2}{N\Upsigma^l+\mk_2^l}\\
&=\frac{\overline{K_{ii}}^2g^4(\mk_1^{l-1})^2}{\overline{K_{ij}^2}\left(g^4(N\Upsigma^{l-1}+\mk_2^{l-1})+r^2(\mk_1^{l-1})^2\right)+\overline{K_{ii}^2}g^4(\mk_1^{l-1})^2}\\
&= 
\frac{(\mk_1^{l-1})^2}{\upgamma_1(N\Upsigma ^{l-1}+\mk_2^{l-1})+\left(\upgamma _1 \frac{r^2}{g^4}+\upgamma_2\right)\left (\mk_1^{l-1}\right )^2}\ ,
\end{aligned}   \label{Iteration of D continuous}
\end{equation}
where $\upgamma_1= \overline{K_{ij}^2}/{\overline{K_{ii}}^2}$, $\upgamma_2= \overline{K_{ii}^2}/\overline{K_{ii}}^2$ and $r=qN^{\frac{3}{2}}$.
When the superscripts of layer index for $K_{ij}$ and $K_{ii}$ are clear, the superscripts are omitted. The iterative equations for computing $\upgamma_1$ and $\upgamma_2$ are 
the same with Eq.~(\ref{g1g2eq}), yet with $\rho=\frac{qN}{g^2}$ for the continuous-weight case.
$Q^l$, $\mk_1^l$, and $\mk_2^l$ can also be computed recursively by following the iterative equations in Sec.~\ref{SM_cont}.

\subsection{Closed-form mean-field iterations for estimating the dimensionality}

For continuous weights, the mean-field iteration of the covariance matrix of activations is given by
\begin{equation}
\Delta^l_{ij} = \sum_{k,k'}w^l_{ik}C^{l-1}_{kk'}w^l_{jk'} , \label{MF Delta binary}
\end{equation}
\begin{equation}
\begin{aligned} 
m_i^l &=  \int Dt\phi \left( \sqrt{\Delta_{ii}^l}t  +\sum_{j=1}^N w_{ij}^l m_j^{l-1} +b_i^l\right) \ ,   
\end{aligned} \label{MF m binary}
\end{equation}
and
\begin{equation}
\begin{aligned}  
C_{ij}^l &= \int D x D y \phi\left(\sqrt{\Delta_{i i}^{l}} x+b_{i}^{l}+\sum_{k} {w}^{l}_{ik} {m}^{l-1}_{k}\right) \times 
\\&\quad \phi\left(\sqrt{\Delta_{j j}^{l}}(\uppsi x+y \sqrt{1-\uppsi^{2}} ) 
+b_{j}^{l}+\sum_{k'}{w}^{l}_{jk'} {m}^{l-1}_{k'} \right)  -m_{i}^{l} m_{j}^{l} \ ,
\end{aligned} \label{MF C binary}
\end{equation}
where $D x={e^{-x^{2} / 2} d x }/{ \sqrt{2 \pi}}$, and $ \uppsi={\Delta_{i j}^{l}}/{\sqrt{\Delta_{i i}^{l} \Delta_{j j}^{l}}}$. 
By taking the covariance matrix $\Lambda$ and zero mean of the data as initial values, we can get the covariance matrix of the activation values of each subsequent 
layer by iterating the above mean-field equations [Eqs.~(\ref{MF Delta binary}), (\ref{MF m binary}), and~(\ref{MF C binary})]. Then according to the formula $D= \left(\operatorname{Tr}\mathbf{C}\right)^2 / \operatorname{Tr}\mathbf{C}^2$, 
the representation dimensionality
of each layer can be calculated.

For binary weights, the equation of the mean-field iteration is just a bit different, given by
\begin{equation}
\Delta_{i j}^{l}=\frac{g^2}{N}\sum_{k,k'}w^l_{ik}C^{l-1}_{kk'}w^l_{jk'} , \label{MF Delta continuous}
\end{equation}
\begin{equation}
\begin{aligned} 
m_i^l &=  \int Dt\phi \left( \sqrt{\Delta_{ii}^l}t  +\frac{g}{\sqrt{N}}\sum_{j=1}^N w_{ij}^l m_j^{l-1} +b_i^l\right) \ ,   
\end{aligned} \label{MF m continuous}
\end{equation}
and
\begin{equation}
\begin{aligned}  
C_{ij}^l  &= \int D x D y \phi\left(\sqrt{\Delta_{i i}^{l}} x+b_{i}^{l}+ \frac{g}{\sqrt{N}} \sum_{k} w_{ik}^l m_{k}^{l-1}\right) \cdot
\\&\quad \phi\left(\sqrt{\Delta_{j j}^{l}}(\uppsi x+y \sqrt{1-\uppsi^{2}} ) 
 +b_{j}^{l}+   \frac{g}{\sqrt{N}} \sum_{k'} w_{jk'}^l m_{k'}^{l-1} \right)  -m_{i}^{l} m_{j}^{l} \ .
\end{aligned} \label{MF C continuous}
\end{equation}
The procedure of estimating the linear dimensionality is similar to that of the continuous-weight case.

\section{Numerical generation of weight-correlated neural networks}

We consider a five-layer fully connected neural network with one input layer and four hidden layers. The number of neurons in each layer is specified by $N$.
The parameters of the network are generated by following the procedure below, and after the initialization, all parameters remain unchanged during the simulation of dimension estimation, and then the result
is averaged over many independent realizations of the same statistics of network parameters.

\subsection{Binary weights}

\begin{figure}
	\centering
	\includegraphics[bb=16 9 174 154,scale=0.8]{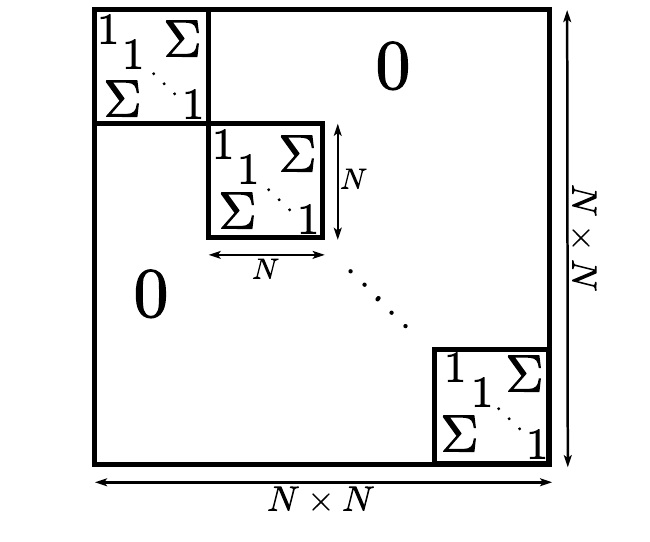}
	\caption{The schematic illustration of the covariance matrix of ${\boldsymbol x}^l$ used to generate correlated binary weights.}
	\label{fig:x covariance}
\end{figure}

The binary weight ($w_{ij}=\pm1$) follows a statistics of zero mean and the covariance specified by
\begin{equation}
\overline{w^l_{i j} w^l_{k s}}=\delta_{j s} {q}+\delta_{i k} \delta_{j s}\left(1-{q}\right)={q} \delta_{j s}\left(1-\delta_{i k}\right)+ \delta_{i k} \delta_{j s}\ .
\end{equation}
Diagonalization of the full covariance matrix of binary weights is challenging. However, no correlation occurs within each RF. Then we can generate the network weights for each diagonal block in Fig.1 (see the main text) independently
by a dichotomized Gaussian (DG) process~\cite{DG-2009}. In the DG process,
the binary weights can be generated by $w_{ij}^l=\operatorname{sign}(x^l_{ij})$, where 
\begin{equation}
\operatorname{sign}(x) = \begin{cases} 1&x\ge 0\\-1 &x<0\end{cases},
\end{equation}
where $x^l_{ij}$ is sampled from a multivariate Gaussian distribution of zero mean (due to $\overline{w_{ij}^l}=0$) and the following covariance, as also shown in a schematic illustration in Fig.~\ref{fig:x covariance},
\begin{equation}
\overline {x^l_{i j} x^l_{k s}}=\delta_{j s} {\Sigma}+\delta_{i k} \delta_{j s}\left(1-{\Sigma}\right)={\Sigma} \delta_{j s}\left(1-\delta_{i k}\right)+ \delta_{i k} \delta_{j s}\ .
\end{equation}
The relation between $q$ and $\Sigma $ can be established by matching the covariance of the DG process with our prescribed correlation level $q$, i.e.,
\begin{equation}
 q=\iint DxDy\operatorname{sign} (x) \operatorname{sign}\left(\Sigma x+\sqrt{1-\Sigma^2}y\right)=\frac{2}{\pi}\arcsin\Sigma.
\end{equation}
Then, we have
\begin{equation}
\Sigma =\sin\frac{\pi q}{2}.
\end{equation}
A sample of the multivariate Gaussian distribution with the $N\times N$ covariance matrix $\boldsymbol{\Sigma}$ (diagonal blocks in Fig.~\ref{fig:x covariance}) can be obtained by first carrying out a Cholesky decomposition of the covariance, i.e.,
$\boldsymbol{\Sigma}=\mathbf{L}\mathbf{L}^{{\rm T}}$, where $\mathbf{L}$ is a lower-triangular matrix. A sample is then obtained as $\mathbf{z}=\mathbf{L}\boldsymbol{\epsilon}$,
where $\boldsymbol{\epsilon}\sim\mathcal{N}(\mathbf{0},\mathbbm{I})$. $\mathbbm{I}$ denotes an identity matrix.
The parameter $b_i^l$ follows $\mathcal{N}(0,\sigma_b)$ independently.

\subsection{Continuous weights}

For the continuous case, the parameter $w_{ij}^l$ follows the Gaussian distribution with zero means and the covariance specified by
\begin{equation}
\overline{w^l_{i j} w^l_{k s}}=\delta_{j s} {q}+\delta_{i k} \delta_{j s}\left(\frac{g^2}{N}-{q}\right)={q} \delta_{j s}\left(1-\delta_{i k}\right)+\frac{g^2}{N} \delta_{i k} \delta_{j s}.
\end{equation}
When generating the weights, we divide $\bw^l$ into $N$ vectors, each of which is defined by 
$[w_{1j}^l,$ $w_{2j}^l, \cdots, w_{Nj}^l]$ ($j=1,2,\cdots,N$). Then all those vectors are independently sampled from a multivariate Gaussian distribution with zero means and the covariance matrix $\boldsymbol{\Sigma}\in \mathbb{R}^{N\times N}$, in which $\boldsymbol{\Sigma}$ is a symmetric matrix with diagonal elements $g^2/N$ and off-diagonal elements $q$. 
The parameter $b_i^l$ follows $\mathcal{N}(0,\sigma_b)$ independently. 

\section{Numerical generation of input data samples following the pre-defined covariance}
\label{samp}
The input dataset for our deep transformation include $100000$ data samples, which are independently sampled from a multivariate Gaussian distribution with zero means and the
covariance matrix specified by $\Lambda=\frac{1}{N}\bxi \bxi^{{\rm T}}$, where $\bxi \in \mathbb{R}^{N\times P}$  and its elements follow a Gaussian distribution $\xi_{ij}\sim\mathcal{N}(0,\sigma^2)$. We define $\alpha =P/N$, and the relation between $\alpha$ and $\tilde{D}$ can
be proved to be $\tilde{D}=\alpha/(1+\alpha)$, as we shall show later.

First, we need to calculate the eigenvalue spectrum of $\Lambda$, namely $\{\lambda_i^\Lambda\}$ by replica trick~\cite{EJ-1976}. The Edwards-Jones formula reads
\begin{equation}
\begin{aligned}
\rho(\lambda)&=\frac{1}{N}\sum_{i=1}^N\delta(\lambda-\lambda_i^\Lambda)\\
&=\frac{1}{N\pi}\lim _{\epsilon \rightarrow 0^{+}} \operatorname{Im}\frac{\partial}{\partial\lambda}\left<\ln\operatorname{det}(\lambda_\epsilon\mathbbm{I}-\Lambda)\right>_{\Lambda}\\
&=\frac{-2}{\pi N} \lim _{\epsilon \rightarrow 0^{+}} \operatorname{Im} \frac{\partial}{\partial \lambda}\langle\ln Z(\lambda)\rangle_\Lambda \ ,
\end{aligned}
\end{equation}
where the Sokhotski-Plemelj identity is used to derive the second equality, and a Gaussian integral representation of the determinant leads to
\begin{equation}
Z(\lambda)=\int_{\mathbb{R}^{N}} d \by \exp \left[-\frac{1}{2} \boldsymbol{y}^{T}\left(\lambda_{\epsilon} \mathbb{I}-\Lambda\right) \boldsymbol{y}\right] \ ,
\end{equation}
where $\lambda_\epsilon = \lambda-\i\epsilon$, and $d\by\defe\prod_i\frac{dy_i}{\sqrt{2\pi}}$.

Hereafter, for simplicity, we consider only the annealed average $\langle Z(\lambda)\rangle_\Lambda$, instead of the more complex quenched one like that analyzed in Ref.~\cite{EJ-1976}.
The result will be cross-checked by numerical experiments. In some cases, the annealed average agrees with the quenched one via replica trick, due to the fact that 
different replicas of the Gaussian variables $\by$ are decoupled~\cite{Stein-1988}. It then proceeds as follows,

\begin{equation}
\begin{aligned}\langle Z(\lambda)\rangle 
&= \int d\by \left\langle\exp\left[ -\frac{1}{2} \boldsymbol{y}^{T}\left(\lambda_{\epsilon} \mathbb{I}-\Lambda\right) \boldsymbol{y}   \right] \right\rangle\\
&=\int d\by \exp\l[-\frac{1}{2} \lame \by^T \by \r]\la   \exp\l[ \frac{1}{2N} \by^T\bxi \bxi ^T\by \r]\ra\\
&=\int d\by \exp\l[-\frac{1}{2} \lame  \sum_{i=1}^N y_i^2  \r]\la   \exp\l[ \frac{1}{2N} \sum_{j=1}^P \l(\sum_{i=1}^N y_i\xi_{ij} \r)^2 \r]\ra\\
&=\int d\by \exp\l[-\frac{1}{2} \lame  \sum_{i=1}^N y_i^2  \r]\la  \prod_{j=1}^P \exp\l[ \frac{1}{2N}  \l(\sum_{i=1}^N y_i\xi_{ij} \r)^2 \r]\ra\\
&\propto \int d\by d\bm \exp\l[-\frac{1}{2} \lame  \sum_{i=1}^N y_i^2-\sum_{j=1}^Pm_j^2\r]\la  \prod_{j=1}^P\prod_{i=1}^N \exp\l[ 2\sqrt{\frac{1}{2N}} m_j  y_i\xi_{ij}  \r]\ra \, ,\end{aligned}
\end{equation}
where $d\bm\defe\prod_i\frac{dm_i}{\sqrt{\pi}}$, and we have used the integral identity for $\bm$:
\begin{equation}
e^{b^2} = \frac{1}{\sqrt{\pi}} \int e^{-x^2+2bx}dx .
\end{equation}
We then calculate the expectation with respect to $\bxi$, and obtain
\begin{equation}
\begin{aligned}
\left\langle  \prod_{j=1}^P \prod_{i=1}^N  \exp\left[ 2\sqrt{\frac{1}{2N}} m_j  y_i \xi_{ij}  \right]\right\rangle_{\bxi} &=  \prod_{j=1}^P\prod_{i=1}^N\la  \exp\l[ 2\sqrt{\frac{1}{2N}} m_j  y_i\xi_{ij}  \r]\ra_{\xi_{ij}}\\&= \prod_{j=1}^P\prod_{i=1}^N  \exp\l[ {\frac{1}{N}} \sigma^2m_j^2  y_i^2  \r] .
\end{aligned}
\end{equation}
Thus,
\begin{equation}
\begin{aligned}\langle Z(\lambda)\rangle &\propto \int d\by d\bm \exp\l[-\frac{1}{2} \lame  \sum_{i=1}^N y_i^2-\sum_{j=1}^Pm_j^2+ {\frac{1}{N}} \sigma^2 \sum_{j=1}^Pm_j^2  \sum_{i=1}^Ny_i^2  \r]\\
&=\int d\by d\bm dqdr\exp\l[-\frac{1}{2} \lame  Nq-Pr + P \sigma^2 rq  \r] \\
&\times \delta\left(Nq-\sum_{i=1}^N y_i^2\right )\delta\left(Pr-\sum_{j=1}^Pm_j^2\right)\\
&\propto \int d\by d\bm dqdrd\qh d\rh \exp\l[-\frac{1}{2} \lame  Nq-Pr + P \sigma^2 rq  \r] \\
&\times \exp\l[Nq\qh- \qh \sum_{i=1}^N y_i^2 \r]\exp\l[Pr\rh-\rh \sum_{j=1}^Pm_j^2\r]\\
&\propto\int  dqdrd\qh d\rh \exp\l[-\frac{1}{2} \lame  Nq-Pr + P \sigma^2 rq  \r] \\
&\times \exp\l[Nq\qh- \frac{N}{2}\ln \qh  \r]\exp\l[Pr\rh-\frac{P}{2}\ln\rh\r],\end{aligned}
\end{equation}
where we have used the following identity:
\begin{equation}
\int dy e^{-ay^2}=\sqrt{\frac{\pi}{a}} =\sqrt{\pi} e^{-\frac{1}{2}\ln a}.
\end{equation}

\begin{figure}
	\centering
	\includegraphics[bb=3 2 392 259,scale=0.8]{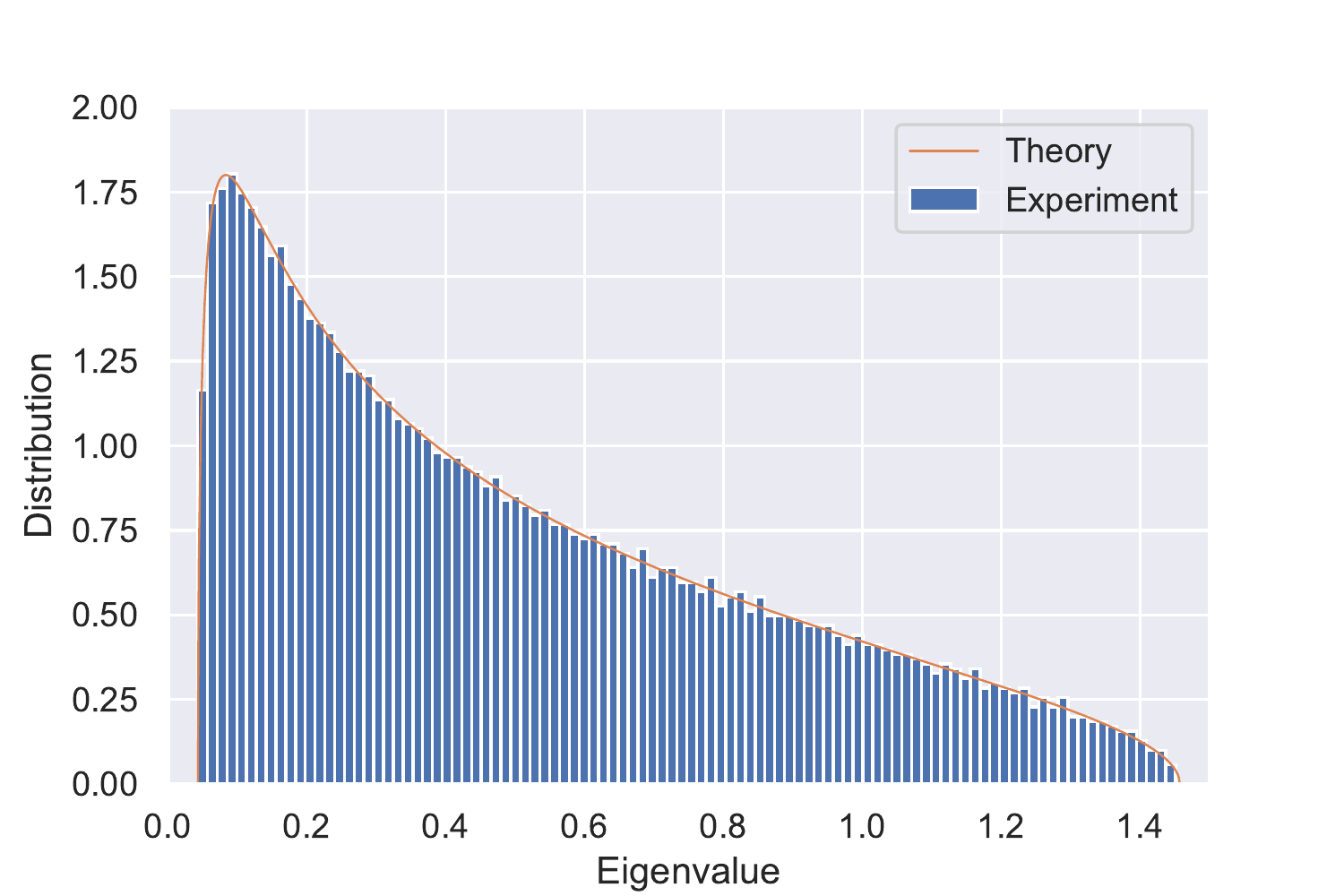}
	\caption{The eigenvalue distribution of the covariance matrix $\Lambda$. Here we set $\alpha=2$, $\sigma=0.5$ and $N=5000$.}
	\label{fig:mp-law}
\end{figure}

We can now write down $\langle Z(\lambda)\rangle $ as
\begin{equation}
\langle Z(\lambda)\rangle =\int dqdrd\qh d\rh \exp\l[-Nf_\lambda(q,r,\qh,\rh) \r],
\end{equation}
where
\begin{equation}
f_\lambda(q,r,\qh,\rh) =\frac{1}{2} \lame  q+\alpha r - \alpha  \sigma^2 rq-q\qh+ \frac{1}{2}\ln \qh -\alpha r\rh+\frac{\alpha}{2}\ln\rh .
\end{equation}
When $N\to\infty$, we can use Laplace's approximation, from which
\begin{equation}
\langle Z(\lambda)\rangle \simeq \exp\l[-Nf_\lambda (q^\star,r^\star,\qh^\star,\rh^\star) \r]\ .
\end{equation}
The stationary point $(q^\star,r^\star,\qh^\star,\rh^\star)$ is computed as
\begin{equation}
\begin{aligned}\frac{ \partial f_\lambda}{\partial q} &=0\Rightarrow \frac{1}{2} \lame  - \alpha  \sigma^2 r-\qh=0\ ,\\
\frac{ \partial f_\lambda}{\partial r} &=0 \Rightarrow \alpha  - \alpha  \sigma^2 q-\alpha \rh=0\ ,\\
\frac{ \partial f_\lambda}{\partial \qh} &=0\Rightarrow q-\frac{1}{2\qh}=0\ , \\ 
\frac{ \partial f_\lambda}{\partial \rh} &=0\Rightarrow\alpha r-\frac{\alpha}{2\rh}=0 \ .
\end{aligned}\label{stationary point}
\end{equation}
Applying the Edwards-Jones formula in the annealed version, we obtain
\begin{equation}
\rho(\lambda )=\frac{-2}{\pi N} \lim _{\varepsilon \rightarrow 0^{+}} \operatorname{Im} \frac{\partial}{\partial\lambda}  \ln \la Z(\lambda)\ra \simeq \frac{-2}{\pi N} \lim _{\varepsilon \rightarrow 0^{+}} \operatorname{Im} \frac{\partial}{\partial \lambda}\left[-N f_\lambda (q^\star,r^\star,\qh^\star,\rh^\star)\right] \ .
\end{equation}
Note that $\frac{\partial f_\lambda}{\partial \lambda}= \frac{1}{2}q^\star$, where $q^\star$ can be obtained by solving the saddle-point equations [Eq.~(\ref{stationary point})].
We finally arrive at
\begin{equation}
\rho(\lambda )\simeq \frac{1}{\pi } \lim _{\epsilon \rightarrow 0^{+}} \operatorname{Im} \left[q^\star\right]= \frac{\sqrt{ (\sigma^2\lambda_+-\lambda)(\lambda- \sigma^2\lambda_-)}}{2\pi\sigma^2 \lambda}\ ,\label{MP law}
\end{equation}
where $\lambda_\pm=(\sqrt{\alpha}\pm 1)^2$ for $\alpha>1$. The comparison between theory and simulation is shown in Fig.~\ref{fig:mp-law}.

\begin{figure}
	\centering
	\includegraphics[bb=14 2 482 216,scale=0.8]{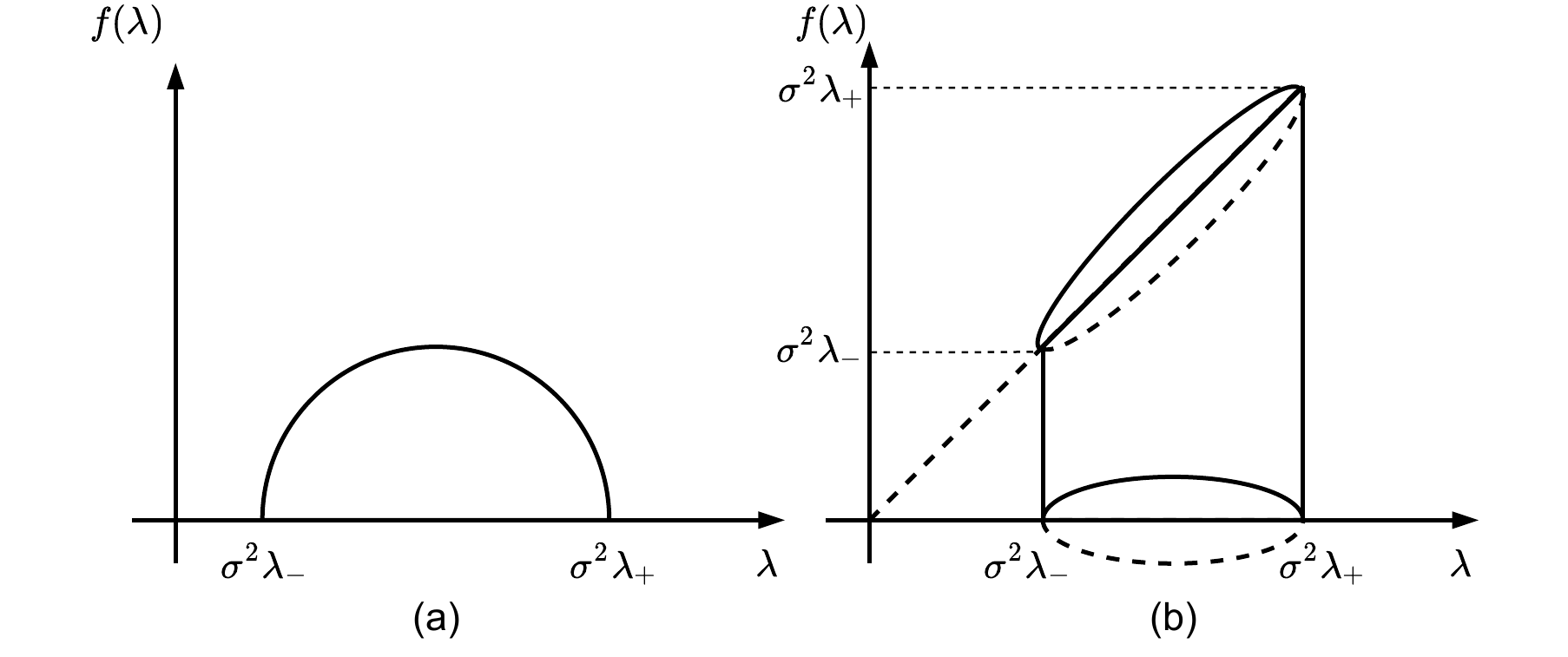}
	\caption{The calculation of spectrum moments by a geometric method. $f(\lambda)$ denotes the integrand of $I_1$ (a) or $I_2$ (b).}
	\label{fig:-integral}
\end{figure}

In the large $N$ limit, the normalized dimensionality of the matrix $\Lambda$ can be written in the form as follows,
\begin{equation}
\tilde{D} = \frac{\left(\sum_i \lambda_i\right)^2}{N\sum_i \lambda_i^2}=\frac{\left( \frac{1}{N}\sum_i \lambda_i\right)^2}{\frac{1}{N}\sum_i \lambda_i^2}=\frac{\left(\int d\lambda \rho(\lambda) \lambda\right)^2}{\int d\lambda \rho(\lambda) \lambda^2} \ .
\end{equation}
The numerator part is given by the first order moment of the eigen-spectrum,
\begin{equation}
\begin{aligned}I_1&\defe \int d\lambda \rho(\lambda) \lambda\\&= \frac{1}{2\pi \sigma^2} \int^{\lambda_+}_{\lambda_-} \sqrt{ (\lambda-\sigma^2\lambda_-)(\sigma^2\lambda_+-\lambda)}d\lambda\\&= \frac{1}{2\pi \sigma^2} \frac{1}{2}\pi \left[ \frac{\sigma^2(\lambda_+-\lambda_-)}{2}\right]^2\\&=\alpha\sigma^2,\end{aligned}
\end{equation}
where we calculate the integral by computing the area of the semicircle shown in Fig.~\ref{fig:-integral} (a).
The denominator part is given by the second moment of the spectrum, computed as 
\begin{equation}
\begin{aligned}I_2 &\defe \int d\lambda \rho(\lambda) \lambda^2  \\&= \frac{1}{2\pi \sigma^2} \int^{\lambda_+}_{\lambda_-}\lambda \sqrt{ (\lambda-\sigma^2\lambda_-)(\sigma^2\lambda_+-\lambda)}d\lambda\\&= \frac{1}{2\pi \sigma^2} \frac{1}{2}\pi \left[ \frac{\sigma^2(\lambda_+-\lambda_-)}{2}\right]^2   \frac{\sigma^2(\lambda_++\lambda_-)}{2}\\&= \sigma^4\alpha(\alpha+1),\end{aligned}
\end{equation}
where the integral here can be transformed to half of the volume of the cylinder shown in Fig.~\ref{fig:-integral} (b)~\cite{ZM-2020}.
Finally, we conclude that
\begin{equation}
\tilde{D} =\frac{I_1^2}{I_2}=\frac{\alpha}{\alpha+1},
\end{equation}
from which the normalized input dimensionality is independent of the input-pattern ($\bxi$) variance $\sigma^2$. This analytic result is confirmed in numerical simulations in the main text.

\section{Proof of $\upgamma_1\le \upgamma_2$}
In this section, we prove the relation $\upgamma_1 \le \upgamma_2$ at an arbitrary layer $l$, 
where $\upgamma_1= \overline{K_{ij}^2}/{\overline{K_{ii}}^2}$ and $\upgamma_2= \overline{K_{ii}^2}/\overline{K_{ii}}^2$. Because the denominators of $\upgamma_1$ and $\upgamma_2$ 
are the same, we just need to prove that $\overline{K_{ij}^2}\le \overline{K_{ii}^2}$ where
\begin{equation}
\begin{aligned}
\overline{K_{ii}^2}&= \left<\left(\phi^{\prime}\left(\sqrt{g^2\mk_1^{l-1}}t+x \sqrt{g^2  Q^{l-1}+\sigma_{b}}\right)\right)^{4}\right>_{x,t} \ , \\
\overline{K_{ij}^2}&=\left< \left<\phi'\left(\sqrt{g^2\mk_1^{l-1}} x+\sqrt{g^2Q^{l-1}}z_1+\sqrt{\sigma_b}u_1\right) \right>^2_x\right.\\
&\left.\times\left<\phi'\left(\sqrt{g^2\mk_1^{l-1}} y+\sqrt{g^2Q^{l-1}}\left(\rho z_1+\sqrt{1-\rho^2}z_2\right)+\sqrt{\sigma_b}u_2\right)\right>^2_y\right> _{z_1,z_2,u_1,u_2} \ .
\end{aligned}
\end{equation}
Note that we add higher-order contributions of $\Delta_{ii}^l$ back into $\overline{K_{ii}^2}$, as mentioned in Sec.~\ref{Expan-corr}.
To proceed, we define $\vartheta_x\defe\sqrt{g^2\mk_1^{l-1}} x+\sqrt{g^2Q^{l-1}}z_1+\sqrt{\sigma_b}u_1$ and $\vartheta_y\defe\sqrt{g^2\mk_1^{l-1}} y+\sqrt{g^2Q^{l-1}}\left(\rho z_1+\sqrt{1-\rho^2}z_2\right)+\sqrt{\sigma_b}u_2$.
Let $\langle\phi'(\vartheta_x)\rangle_x^2=\Phi_1$ and $\langle\phi'(\vartheta_y)\rangle_y^2=\Phi_2$. According to the Cauchy-Schwartz inequality, we have 
\begin{equation}
 \overline{\Phi_1\Phi_2}^2\leq\overline{\Phi_1^2}\cdot\overline{\Phi_2^2},
\end{equation}
where $\overline{\ \cdot\ }$ denotes the quenched disorder average as before. It is then easy to show that $\overline{\Phi_2^2}=\overline{\Phi_1^2}=\overline{K_{ii}^2}$. We finally conclude that
$\upgamma_1\leq\upgamma_2$.

\section{Deep neural networks trained with Hebbian learning rules}
To verify the revealed principles in this paper, we perform an on-line training of a layered neural network using the same transfer function, by applying the well-known Hebbian rule.
We use the synthetic dataset generated in Sec.~\ref{samp}, containing $10\ 000$ input samples ($\alpha=2$). These data samples are then sequentially
shown to the input layer of the deep neural network, and then learned layer by layer. We use the synaptic rescaling to control the synaptic strength, like
$w_{ij}(t)\leftarrow\frac{gw_{ij}(t)}{\sqrt{\sum_kw_{ik}^2(t)}}$, where $w_{ij}(t)$ is updated by the following regularized Hebbian rule:
\begin{equation}\label{Hebb}
 w_{ij}^l(t)=w_{ij}^l(t-1)+\eta\left[h_{i}^lh_j^{l-1}-\kappa_c\left(\sum_{i':i'\neq i}w_{i'j}^l\right)
 \left(\sum_{k\neq k'}^Nw_{kj}^lw_{k'j}^l-\sqrt{N}r\right)\right],
 \end{equation}
where $t$ denotes the learning step, $N$ denotes the receptive field size of the hidden neuron at the $l$-th layer, $\eta$ is the learning rate, and
$\kappa_c$ enforces
the synaptic-correlation constraint, inspired by our theory. The last term in Eq.~(\ref{Hebb}) can be derived as the gradient descent of the correlation-constraint
objective:
\begin{equation}
 \Phi_{c}(\mathbf{w})=\frac{\kappa_c}{2}\sum_j\left[\sum_{i\neq i'}^Nw_{ij}w_{i'j}-\sqrt{N}r\right]^2,
\end{equation}
where the synaptic-correlation scaling derived in our paper is used for learning the continuous weights. The synaptic rescaling operation together with
the synaptic-correlation constraint encourages synapses to compete with each other to encode input features during learning~\cite{MacKay-1994,Seung-2017}.
\begin{figure}
	\centering
	\includegraphics[bb=2 5 380 311,scale=0.6]{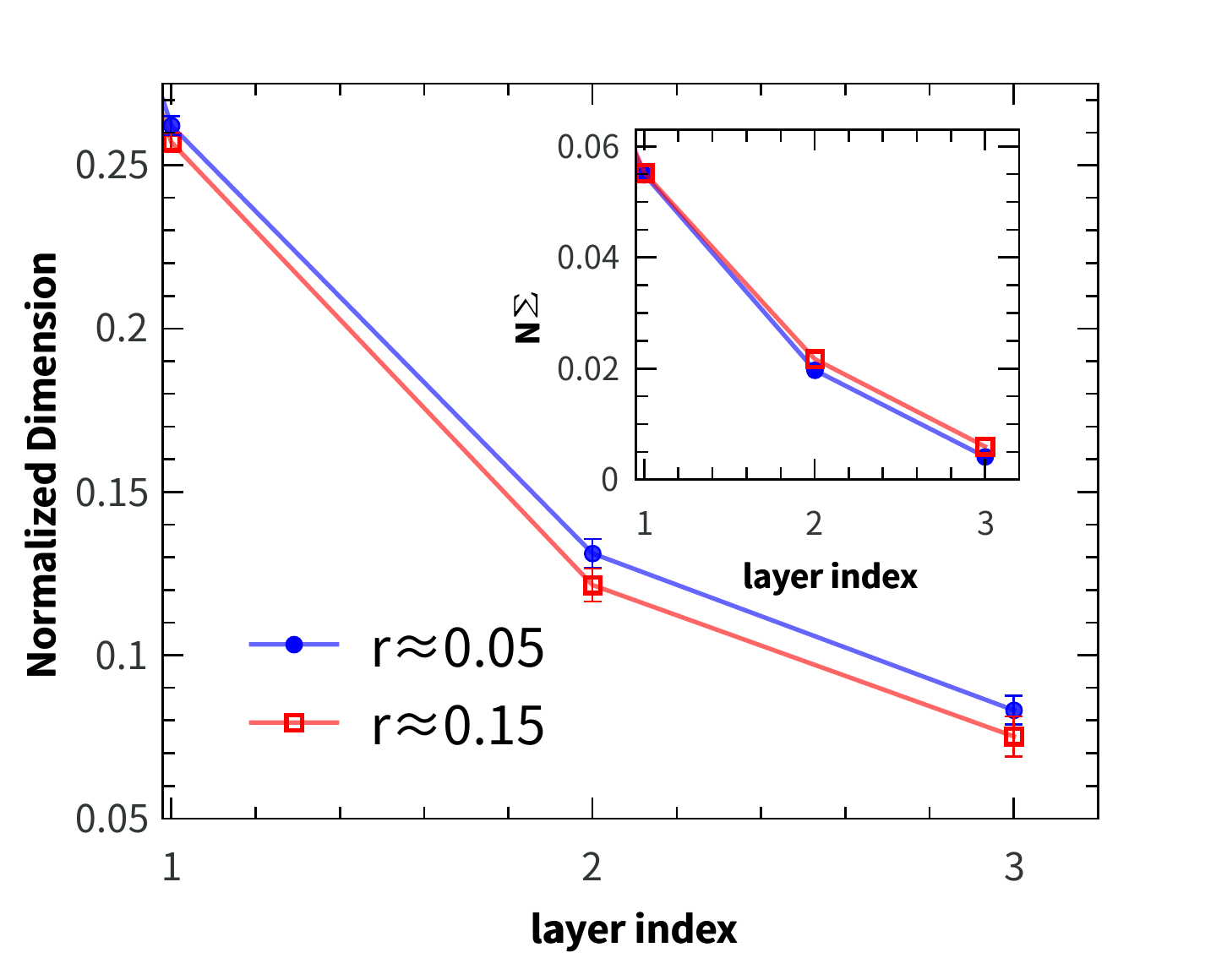}
	\caption{Dimension reduction and decorrelation in deep Hebbian learning with $N=100$. $\eta=0.0001$, $\kappa_c=0.5$, $g=0.5$, and $10\ 000$ training examples ($\alpha=2$) are sequentially (on line)
	shown to the input layer of deep networks. The result is averaged over ten random realizations.}
	\label{Hebb-res}
\end{figure}

As shown in Fig.~\ref{Hebb-res}, the regularized Hebbian learning rule is able to reduce the dimensionality of hidden representations, while decorrelating the representations as well.
The qualitative behavior of the on-line trained systems coincides with the theoretical predictions of our model about roles of synaptic correlations. Therefore, it is promising to design unsupervised/supervised learning
algorithms that can control the synaptic correlations in future works, e.g., in sensory perception of real-world datasets.



\end{document}